\documentclass[]{opendatalab}
\usepackage[table]{xcolor}
\usepackage{longtable}
\usepackage{listings}
\usepackage{marvosym}
\usepackage[numbers]{natbib}
\usepackage{tabularx} 
\usepackage{minted} 
\usepackage{multirow}
\usepackage{booktabs}
\usepackage{tablefootnote}
\usepackage{wrapfig}
\usepackage{enumitem}
\usepackage[T1]{fontenc}
\definecolor{codegreen}{rgb}{0,0.6,0}
\definecolor{codegray}{rgb}{0.5,0.5,0.5}
\definecolor{codepurple}{rgb}{0.58,0,0.82}
\definecolor{backcolour}{rgb}{0.95,0.95,0.92}
\definecolor{promptcolor}{HTML}{D1D0F2}
\definecolor{promptcolorheader}{HTML}{bdbcec}
\definecolor{lightgreen}{rgb}{0.85,0.95,0.85}
\newcommand{\promptbox}[2]{
\begin{tcolorbox}[
top=0.3em,bottom=0.3em,left=0.5em,right=0.5em,
toptitle=0.3em,bottomtitle=0.2em,boxsep=0pt,
colframe=promptcolorheader,colback=promptcolor!50,boxrule=0.5pt,
]
\footnotesize
\end{tcolorbox}
}
\lstdefinestyle{mystyle}{
    backgroundcolor=\color{backcolour},   
    commentstyle=\color{codegreen},
    keywordstyle=\color{magenta},
    numberstyle=\tiny\color{codegray},
    stringstyle=\color{codepurple},
    basicstyle=\ttfamily\footnotesize,
    breakatwhitespace=false,         
    breaklines=true,                 
    captionpos=b,                    
    keepspaces=true,                 
    numbers=left,                    
    numbersep=5pt,                  
    showspaces=false,                
    showstringspaces=false,
    showtabs=false,                  
    tabsize=2
}

\lstdefinelanguage{json}{
    basicstyle=\normalfont\ttfamily,
    numbers=left,
    numberstyle=\scriptsize,
    breaklines=true,
    frame=lines,
    backgroundcolor=\color{gray!10},
    showstringspaces=false,
    string=[db]{"},
    stringstyle=\color{green!50!black},
    morestring=[s][\color{black}]{\ \ "}{":},
    keywordstyle=\color{blue},
    keywords={true,false,null},
    literate=
     *{0}{{{\color{red}0}}}{1}
      {1}{{{\color{red}1}}}{1}
      {2}{{{\color{red}2}}}{1}
      {3}{{{\color{red}3}}}{1}
      {4}{{{\color{red}4}}}{1}
      {5}{{{\color{red}5}}}{1}
      {6}{{{\color{red}6}}}{1}
      {7}{{{\color{red}7}}}{1}
      {8}{{{\color{red}8}}}{1}
      {9}{{{\color{red}9}}}{1}
      {.}{{{\color{red}.}}}{1}
      {:}{{{\color{gray}{:}}}}{1}
      {,}{{{\color{gray}{,}}}}{1}
      {\{}{{{\color{gray}{\{}}}}{1}
      {\}}{{{\color{gray}{\}}}}}{1}
      {[}{{{\color{gray}{[}}}}{1}
      {]}{{{\color{gray}{]}}}}{1},
}

\newcommand{\runningtitle}{}
\newcommand{\setrunningtitle}[1]{\renewcommand{\runningtitle}{#1}}
\title{AICC: Parse HTML Finer, Make Models Better —— \\ A 7.3T AI-Ready Corpus Built by a Model-Based HTML Parser}

\setrunningtitle{AICC: A 7.3T AI-Ready Corpus Built by a Model-Based HTML Parser}
\author[*]{Ren Ma}
\author[*\ \dagger]{Jiantao Qiu}
\author[*\ \dagger]{Chao Xu}
\author[]{Pei Chu}
\author[]{Kaiwen Liu}
\author[]{Pengli Ren}
\author[]{Yuan Qu}
\author[]{Jiahui Peng}
\author[]{Linfeng Hou}
\author[]{Mengjie Liu}
\author[]{Lindong Lu}
\author[]{Wenchang Ning}
\author[]{Jia Yu}
\author[]{Rui Min}
\author[]{Jin Shi}
\author[]{Haojiong Chen}
\author[]{Peng Zhang}
\author[]{Wenjian Zhang}
\author[]{Qian Jiang}
\author[]{Zengjie Hu}
\author[]{Guoqiang Yang}
\author[]{Zhenxiang Li}
\author[]{Fukai Shang}
\author[]{Runyuan Ma}
\author[]{Chenlin Su}
\author[]{Zhongying Tu}
\author[]{Wentao Zhang}
\author[]{Dahua Lin}
\author[\textrm{\Letter}]{Conghui He}

\affiliation{Shanghai Artificial Intelligence Laboratory}

\abstract{
While web data quality is crucial for large language models, most curation efforts focus on filtering and deduplication, treating HTML-to-text extraction as a fixed pre-processing step. Existing web corpora rely on heuristic-based extractors like Trafilatura, which struggle to preserve document structure and frequently corrupt structured elements such as formulas, code blocks, and tables. We hypothesize that improving extraction quality can be as impactful as aggressive filtering strategies for downstream performance.

We introduce \textbf{MinerU-HTML}, a novel extraction pipeline that reformulates content extraction as a sequence labeling problem solved by a 0.6B-parameter language model. Unlike text-density heuristics, MinerU-HTML leverages semantic understanding and employs a two-stage formatting pipeline that explicitly categorizes semantic elements before converting to Markdown. Crucially, its model-based approach is inherently scalable, whereas heuristic methods offer limited improvement pathways. On WebMainBench, our benchmark of 7,887 annotated web pages, MinerU-HTML achieves 81.82\% ROUGE-N F1 compared to Trafilatura's 63.58\%, with exceptional structured element preservation (90.93\% for code blocks, 93.99\% for formulas).

Using MinerU-HTML, we construct \textbf{AICC (AI-ready Common Crawl)}, a 7.3-trillion token multilingual corpus from two Common Crawl snapshots. In controlled pretraining experiments where AICC and Trafilatura-extracted TfCC undergo identical filtering, models trained on AICC (62B tokens) achieve 50.82\% average accuracy across 13 benchmarks, outperforming TfCC by 1.08pp—providing direct evidence that extraction quality significantly impacts model capabilities. AICC also surpasses RefinedWeb and FineWeb on key benchmarks. We publicly release WebMainBench, MinerU-HTML, and AICC, demonstrating that HTML extraction is a critical, often underestimated component of web corpus construction.
}

\metadata[Website:]{\url{https://opendatalab.com/ai-ready/AICC}}
\correspondence{Conghui He, \email{heconghui@pjlab.org.cn}}
\StarMark{Contributed equally}
\DaggerMark{Project leader}

\begin{document}

\maketitle

\newpage
\section{Introduction}

The remarkable capabilities of modern large language models (LLMs) are built upon massive-scale pretraining on diverse text corpora~\citep{brown2020language,llama,touvron2023llama2,dubey2024llama}. As models scale to hundreds of billions of parameters and training extends to trillions of tokens, the quality and quantity of pretraining data have become critical determinants of model performance~\citep{hoffmann2022training}. Among the various data sources available, web text has emerged as the dominant component due to its unparalleled scale and diversity. Common Crawl~\citep{commoncrawl}, a continuously updated public repository of web snapshots containing petabytes of HTML documents, has become the de facto foundation for constructing large-scale pretraining corpora.

However, transforming raw Common Crawl data into effective training material is far from trivial. Recent efforts have demonstrated substantial improvements in downstream model performance through sophisticated data curation strategies. RefinedWeb~\citep{refinedweb} showed that extensively filtered and deduplicated web data alone can outperform curated corpora that mix web text with books and technical documents. FineWeb~\citep{fineweb} introduced careful ablation studies of filtering and deduplication strategies, producing a 15-trillion token corpus that yields better-performing models than other public web datasets. DCLM~\citep{DCLM2024} demonstrated that model-based quality filtering can dramatically improve benchmark performance, while Nemotron-CC~\citep{nemotron2024} explored the trade-off between aggressive filtering and maintaining sufficient data quantity for long-horizon training.

Despite this progress in data curation, a critical component of the pipeline has received comparatively little attention: \textbf{HTML-to-text extraction}. Before any filtering or deduplication can be applied, raw HTML documents must first be converted to structured text formats. This extraction step faces significant challenges: HTML is primarily designed for rendering visual layouts, not for conveying semantic content. Web pages are riddled with navigation menus, advertisements, sidebars, footers, and other boilerplate elements that must be separated from the main content. Furthermore, structured elements such as mathematical formulas, code blocks, and tables—which are crucial for technical and educational content—are often corrupted or lost entirely during extraction.

Existing web corpora have largely relied on heuristic-based extraction tools: RefinedWeb and FineWeb use Trafilatura~\citep{barbaresi2021Trafilatura}, while DCLM and Dolma~\citep{dolma} use Resiliparse~\citep{Resiliparse}. Though DCLM found that both extractors significantly outperform Common Crawl's WET files (pre-extracted text), improving MMLU scores by 2.5+ points, these tools remain fundamentally limited by their reliance on text density heuristics and hand-crafted DOM traversal rules. Such approaches struggle with non-standard layouts, fail to preserve document structure and coherence, and frequently corrupt structured elements. For instance, mathematical formulas rendered via MathJax are often reduced to raw LaTeX commands or stripped entirely; code blocks lose indentation and syntax highlighting markers; complex tables are flattened into unstructured text sequences. More critically, most recent work has treated extraction as a fixed pre-processing step, focusing optimization efforts exclusively on downstream filtering and deduplication.

We hypothesize that \textbf{improving HTML extraction quality can be as impactful as aggressive filtering strategies}, yet requires fundamentally different methods. In this work, we introduce \textbf{AICC (AI-ready Common Crawl)}, a large-scale pretraining corpus constructed by applying semantic-aware HTML extraction to Common Crawl. At the core of AICC is \textbf{MinerU-HTML}, a novel two-stage extraction pipeline that reformulates content extraction as a sequence labeling problem solved by a compact 0.6B-parameter language model. Unlike heuristic methods, MinerU-HTML understands semantic context and preserves document structure, enabling high-fidelity extraction of main content while maintaining formulas, code blocks, and tables.

Our comprehensive evaluation demonstrates that extraction quality significantly impacts downstream model capabilities:

\begin{enumerate} [leftmargin=*,topsep=0pt,itemsep=2pt]
\item \textbf{Extraction Quality Matters}: On WebMainBench, our newly constructed benchmark of 7,887 annotated web pages, MinerU-HTML achieves a ROUGE-N F1 score of 0.8182, substantially outperforming Trafilatura (0.6358). For structured element preservation, MinerU-HTML achieves 0.9093 edit similarity for code blocks and 0.9399 for formulas, compared to 0.1305 and 0.6107 for Trafilatura.

\item \textbf{Corpus-Level Quality Improvement}: A direct comparison on 10,000 document pairs from Common Crawl, evaluated via an LLM-as-judge, shows AICC is preferred over its Trafilatura-based counterpart (TfCC) with a 72.0\% win rate. This confirms that MinerU-HTML's extraction improvements on the benchmark generalize to the full corpus.

\item \textbf{Extraction Impacts Downstream Performance}: Models pretrained on AICC (62B tokens) achieve 50.82\% average accuracy across 13 diverse benchmarks, outperforming models trained on TfCC (49.74\%), RefinedWeb (49.13\%), and even FineWeb (49.61\%). The 1.08pp improvement over TfCC—where both datasets undergo identical filtering and deduplication, differing only in extraction method—provides direct evidence that extraction quality significantly affects model capabilities.

\item \textbf{Structure Preservation Drives Gains}: The improvements are consistent across all task categories, validating our hypothesis that preserving document structure and narrative coherence during extraction enhances the learning of contextual understanding and long-range dependencies. Notably, AICC significantly outperforms FineWeb on reading comprehension tasks (42.37\% vs. 36.68\%, +5.69pp).
\end{enumerate}

Our work makes the following contributions: (1) We introduce MinerU-HTML, a semantic-aware HTML extraction pipeline that significantly outperforms existing heuristic methods on both content fidelity and structured element preservation. Critically, MinerU-HTML's model-based approach offers inherent scalability advantages over rule-based extractors, as its performance can improve with more data and stronger base models (Figure~\ref{fig:MinerU-HTML_iterate}). (2) We construct and publicly release WebMainBench, a comprehensive benchmark for evaluating main content extraction and structured element preservation with 7,887 annotated pages. (3) We release AICC, a large-scale pretraining corpus demonstrating that high-quality extraction can rival or exceed the benefits of aggressive filtering strategies for downstream model performance. (4) We provide extensive pretraining experiments isolating the contribution of extraction quality, shifting attention to this critical but often overlooked component of web data curation.

\section{MinerU-HTML: HTML Parsing for AI-ready CC}
We present \textbf{MinerU-HTML}, a novel two-stage pipeline for extracting high-quality content from raw HTML documents and converting it into AI-ready formats. In the first stage, MinerU-HTML extracts the main content from raw HTML to produce what we term \textit{Main-HTML}—a cleaned subset of the original document that retains only content-bearing elements. In the second stage, we transform this Main-HTML into structured, AI-ready formats such as Markdown for downstream language model training.

\subsection{Main-HTML Extraction}

MinerU-HTML's Main-HTML extraction methodology is designed with scalability as a central consideration. We begin by developing a robust extraction pipeline for individual HTML documents that leverages a compact language model to perform content classification with high accuracy and reliability. This single-document pipeline forms the foundation of MinerU-HTML. To scale this method to Common Crawl's hundreds of billions of documents, we then introduce a template-aware optimization strategy that exploits the structural regularity of web pages. By clustering pages generated from similar templates and distilling the language model's decisions into reusable extraction rules, we achieve web-scale processing efficiency while preserving the quality of the core extraction pipeline. We first describe the single-document extraction framework (\ref{sec:core_extraction}), then present the scaling strategy (\ref{sec:scaling}).

\subsubsection{Core Extraction Pipeline}
\label{sec:core_extraction}
\begin{figure}[t!]
    \centering
    \includegraphics[width=1\linewidth]{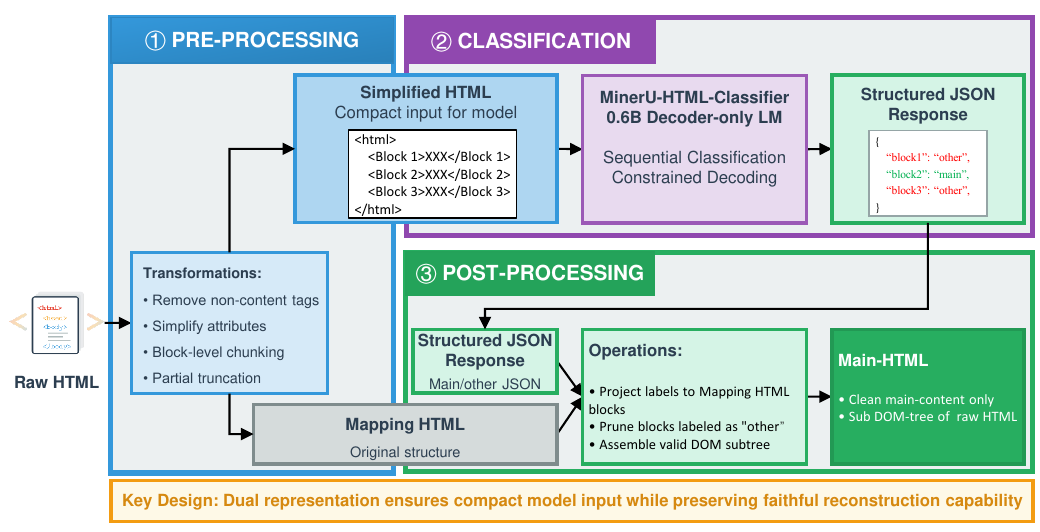}
    \vspace{-0.5cm}
    \caption{\textbf{Overview of the MinerU-HTML Core Extraction Pipeline}. The pipeline consists of three stages: (1) \textbf{Pre-processing}: Raw HTML is transformed into two parallel representations—\textbf{Simplified HTML} (streamlined input for the model with reduced tokens) and \textbf{Mapping HTML} (preserving original structure for faithful reconstruction). (2) \textbf{Content Classification}: MinerU-HTML-Classifier (0.6B parameter LM) performs sequential block classification on the simplified input, with a custom logits processor implementing constrained decoding to ensure structured JSON output without hallucination. (3) \textbf{Post-processing}: Predicted labels ("main" or "other") are used to select corresponding blocks from the Mapping HTML, yielding the final \textbf{Main-HTML} as a valid DOM subtree of the original document.}
    \label{fig:overview}
\end{figure}

MinerU-HTML transforms raw HTML documents into clean Main-HTML through a three-stage pipeline, as illustrated in Figure~\ref{fig:overview}. The architecture is designed to address two key challenges: (1) reducing the computational burden of processing lengthy HTML markup, and (2) ensuring that the extracted content remains faithful to the original document without hallucination.

The pipeline operates as follows. In the \textit{pre-processing} stage, the input HTML is partitioned into semantic blocks, generating two synchronized representations: \textit{Simplified HTML}, which strips away rendering-oriented markup to create compact input for the language model, and \textit{Mapping HTML}, which preserves the original block structure to enable faithful reconstruction. In the \textit{Content Classification} stage, MinerU-HTML-Classifier, a compact 0.6B-parameter language model, processes the Simplified HTML and classifies each block as either main content or boilerplate, with output constrained by a custom logits processor to ensure structured formatting. Finally, in the \textit{post-processing} stage, the predicted labels are projected back onto the Mapping HTML, non-content blocks are pruned, and the resulting Main-HTML is assembled as a valid subtree of the original DOM.

This architecture reformulates content extraction as a sequence labeling problem, offering significant computational advantages over generative approaches. We detail each component below.

\paragraph{Pre-processing and Post-processing}
\label{sec:pre-processing}
The pre-processing stage serves as the foundation of our approach, addressing the fundamental incompatibility between HTML's rendering-oriented structure and the input requirements of language models. Raw HTML documents contain extensive markup designed for visual presentation rather than semantic interpretation, resulting in prohibitively long sequences when directly fed to a model. To overcome this limitation, our pre-processing module implements a systematic transformation strategy that combines simplification and chunking operations to reduce sequence length while preserving semantic information.

The pre-processing pipeline executes four sequential transformations. \textbf{(1) Non-content tag removal}: We eliminate elements such as \texttt{<style>}, \texttt{<script>}, \texttt{<header>}, and \texttt{<aside>} that rarely contribute to main content. \textbf{(2) Attribute simplification}: We retain only the \texttt{class} and \texttt{id} attributes while pruning all others, as these two attributes provide the most salient semantic indicators for content block differentiation. \textbf{(3) Block-level chunking}: The document is segmented at elements that induce line breaks during rendering. This approach preserves the structural integrity of cohesive units such as tables (\texttt{<table>}) and lists (\texttt{<ul>}) by treating them as atomic, indivisible blocks. To accommodate the widespread practice of using tables for page layout, we apply adaptive heuristics that permit selective splitting when structural analysis indicates layout rather than tabular data. \textbf{(4) Partial content truncation}: To manage excessively long individual blocks—such as tables with numerous cells, lists with extensive items, or verbose paragraphs—we retain only representative subsets. For instance, we preserve a subset of table cells or truncate paragraphs to their initial 200 characters. Our empirical validation demonstrates that such partial representations maintain sufficient information for accurate classification while substantially reducing input length.

Following pre-processing, the document is transformed into a sequence of simplified blocks, denoted as $\mathbf{x}=[x_{1},x_{2},\dots,x_{n}]$. In parallel, the Mapping HTML is generated by applying only the block-level chunking to the original, unmodified HTML, ensuring that the final output constitutes a valid DOM subtree. The post-processing module then leverages the model's classification predictions to select content-bearing blocks from the Mapping HTML and assembles them into the final Main-HTML.

\paragraph{Content Classification}
\label{sec:content_classification}
The pre-processing transformations described above enable us to reformulate content extraction as a well-defined \textbf{sequence labeling problem}, analogous to tasks such as named entity recognition or part-of-speech tagging. This formulation provides both theoretical clarity and practical advantages for model training and inference.

Formally, let an HTML document be represented as a sequence of $n$ simplified blocks after pre-processing, denoted as $X = [x_1, x_2, \dots, x_n]$. Each block $x_i$ is associated with a ground-truth binary label $y_i \in \{0, 1\}$, where $y_i = 1$ indicates that block $x_i$ belongs to the main content, and $y_i = 0$ denotes boilerplate or auxiliary content. The objective is to train a language model (i.e., MinerU-HTML-Classifier) $f_\theta$ parameterized by $\theta$ that maps the input sequence to a predicted label sequence:
\begin{equation}
Y_{\text{pred}} = f_\theta(X) = [y'_1, y'_2, \dots, y'_n], \quad y'_i \in \{0, 1\}
\end{equation}
For details of the MinerU-HTML-Classifier, please refer to Appendix \ref{appdx:MinerU-HTML_classifier}.
The post-processing module subsequently utilizes $Y_{\text{pred}}$ to select the corresponding blocks from the Mapping HTML and construct the final Main-HTML output.

This sequence labeling formulation offers several advantages. First, the pre-processing transformations substantially reduce the token load imposed on the model, enabling efficient processing of web documents. Second, by constraining the task to discrete block classification rather than free-form text generation, we limit the output to a compact sequence of binary labels. This design inherently eliminates hallucination risks, as the model can only select from existing content rather than generate novel text, thereby guaranteeing that the extracted content constitutes a faithful subset of the original document.

While the sequence labeling framework provides a strong foundation, standard language model decoding could still produce malformed outputs or introduce spurious tokens. To ensure perfect adherence to the required output format, we employ a constrained decoding mechanism that we describe next.

\paragraph{Constrained Decoding}
\label{sec:constrained_decoding}
To guarantee valid output formatting and completely eliminate the possibility of hallucination or format errors, we implement a custom logits processor that operates as a deterministic finite state machine (FSM) during decoding. The FSM enforces strict control over the generation of structured output, which follows a JSON-like format (e.g., \texttt{\{"1": "main", "2": "other", ...\}}). The processor deterministically manages all syntactic tokens—including braces, quotes, colons, and numeric keys—by masking the logits at each decoding step to permit only valid continuations according to the current state.

The model is granted probabilistic choice exclusively at the critical decision points of block classification, where the vocabulary is restricted to only two tokens: \texttt{"main"} and \texttt{"other"}. This constraint effectively transforms the task into a sequence of binary classification decisions, each made with high confidence over a minimal vocabulary. By eliminating all degrees of freedom except for the semantic classification choices, this mechanism guarantees syntactically valid output and fundamentally precludes both format errors and hallucinated content. 

\begin{figure}[t!]
    \centering
    \includegraphics[width=0.85\linewidth]{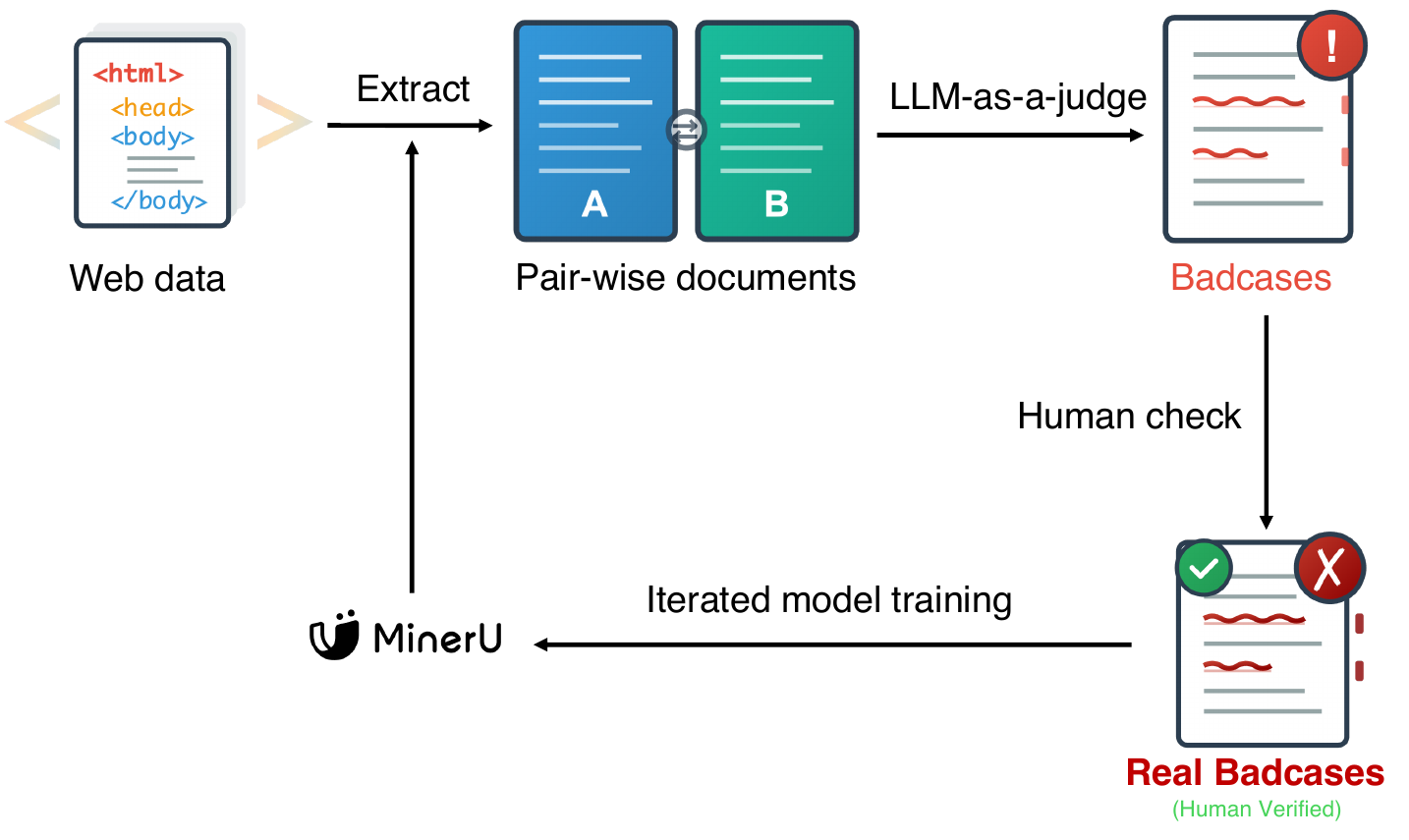}
    \vspace{-0.3cm}
    \caption{\textbf{Iterative improvement pathways for MinerU-HTML}. MinerU-HTML follows a virtuous cycle: the model-based extractor can be systematically improved by collecting more training data (including failure cases), retraining on expanded datasets, and leveraging advances in base model capabilities. This makes MinerU-HTML's approach inherently more scalable and future-proof as language models continue to advance.}
    \label{fig:MinerU-HTML_iterate}
\end{figure}

\paragraph{Scalability Advantages of Model-Based Extraction.} Crucially, MinerU-HTML's approach—grounded in the semantic understanding ability of MinerU-HTML-Classifier is inherently scalable: performance can improve with more training data and stronger base models (Figure~\ref{fig:MinerU-HTML_iterate}). In contrast, purely heuristic methods like Trafilatura and Resiliparse offer limited improvement pathways, as rule updates designed to address specific failure modes often fail to generalize and may introduce conflicts with existing heuristics. This fundamental advantage makes MinerU-HTML a more future-proof solution as language models continue to advance.

\subsubsection{Scaling to Common Crawl}
\label{sec:scaling}
The three-stage pipeline described in Section~\ref{sec:core_extraction} provides an effective solution for extracting content from individual HTML documents. However, directly applying this approach to Common Crawl—which contains hundreds of billions of documents—would require running GPU-based language model inference on each page, resulting in prohibitive computational costs. To achieve web-scale processing, we introduce a template-aware optimization strategy that significantly reduces the number of pages requiring language model inference while maintaining extraction quality.

\paragraph{Template-based generalization.} Our approach exploits the structural regularity inherent in web content: most websites generate pages from shared templates, and pages instantiated from the same template exhibit nearly identical DOM structures. Rather than treating each page independently, we cluster structurally similar pages and apply MinerU-HTML-Classifier (Section~\ref{sec:core_extraction}) to only a small representative subset. We then distill MinerU-HTML-Classifier's extraction decisions into explicit XPath-based rules and propagate these rules to all remaining pages in the cluster using efficient CPU-based processing.

This template-based generalization strategy preserves the high-quality extraction provided by MinerU-HTML-Classifier while dramatically reducing computational requirements. Specifically, we cluster pages at the subdomain level to maximize structural homogeneity, then execute the following procedure for each cluster:

\begin{enumerate}
  \item \textbf{Representative selection}: For each cluster, we select a representative page that maximizes structural coverage and diversity based on tag and attribute variety, DOM depth, and DOM width.
  
  \item \textbf{Full pipeline execution}: We apply the complete three-stage pipeline (Section~\ref{sec:core_extraction}) to the representative page. MinerU-HTML-Classifier processes the Simplified HTML and produces block-level classification labels via constrained decoding, which are used to construct the Main-HTML.
  
  \item \textbf{Rule synthesis}: We analyze the model's classification decisions by mapping the labeled blocks back to their positions in the original DOM structure. From these decisions, we derive generalized extraction rules expressed as XPath or CSS selectors combined with retain/prune operations.
  
  \item \textbf{Rule propagation}: The synthesized rules are applied to all other pages within the cluster. Since rule application requires only DOM traversal and filtering, this step can be executed efficiently on CPUs without requiring GPU resources or language model inference.
\end{enumerate}

The clustering-based optimization delivers significant computational efficiency gains. In practice, a typical subdomain cluster encompasses thousands of pages with similar structural patterns, yet requires language model inference for only a single representative page. All remaining pages in the cluster are processed through efficient rule-based extraction methods. Our empirical analysis of Common Crawl's 300 billion HTML documents reveals that this strategy produces approximately 1.2 billion distinct clusters, thereby requiring GPU inference for only \textbf{~0.4\%} of total pages while preserving extraction quality equivalent to individual page processing.

\subsection{Document Formatting}
The MinerU-HTML Main-HTML extraction pipeline described in the previous section produces cleaned HTML documents that contain only content-bearing elements. However, HTML remains a markup language optimized for rendering rather than for training language models. To maximize the utility of the extracted content for downstream AI applications, we require a format that is both semantically rich and compact. This section presents MinerU-HTML's two-stage formatting pipeline that transforms Main-HTML into AI-ready representations.

We adopt a two-stage conversion strategy with an intermediate representation. First, we parse Main-HTML into a structured \textit{content list}—a JSON-based representation that explicitly categorizes each semantic unit by type (e.g., title, paragraph, code block, table, formula). This intermediate representation enables flexible filtering and format-specific rendering. Second, we convert the content list into the target format, with Markdown being our primary output for language model training due to its balance of expressiveness and simplicity.

\subsubsection{HTML to Content List Conversion}
The first stage transforms Main-HTML into a structured content list that explicitly represents the semantic type of each content element.

\paragraph{Semantic type classification.} Despite the diverse ways HTML elements can be combined, the semantic content of web pages can be systematically categorized into a finite set of types: titles, text paragraphs, inline formulas, display formulas, inline code, code blocks, tables, lists, images, videos, and audio. We implement specialized detectors for each semantic type, processing the Main-HTML through these detectors sequentially to construct a JSON-based content list. Each semantic type is encoded using a dedicated JSON schema. An example of the content list is represented in Figure~\ref{fig:content_list_sample}.

The conversion process incorporates specialized handling for several element types that require sophisticated recognition:
\begin{enumerate}
  \item \textbf{Code blocks}: We aggregate spatially proximate code-related HTML tags to form contiguous code blocks, then apply heuristic pattern matching to infer the programming language based on syntax characteristics.
  
  \item \textbf{Tables}: We distinguish structurally simple tables (expressible in Markdown syntax) from complex tables containing row/column spans or nested structures. Simple tables are converted to Markdown format, while complex tables are preserved as HTML to maintain structural fidelity. Nested tables are linearized into sequential paragraphs.
  
  \item \textbf{Mathematical formulas}: We employ a hybrid extraction strategy combining a renderer-agnostic parser with specialized tag-based extractors. The parser accommodates common rendering engines including MathJax and KaTeX. Extracted formulas are classified as inline (delimited by single dollar signs) or display mode (delimited by double dollar signs).
\end{enumerate}

\paragraph{Advantages of intermediate representation.} The content list serves as a versatile intermediate representation that provides three key benefits. First, it enables \textit{selective filtering} by semantic type, allowing corpus curation based on specific content characteristics (e.g., isolating code-rich or formula-rich pages). Second, it facilitates \textit{format-specific pruning}, where downstream conversion can omit particular element types (e.g., excluding images, audio, or video for language model training). Third, it provides \textit{unified representation} across heterogeneous document sources, including PDFs, DOC files, PowerPoint presentations, EPUB books, and web pages, thereby enabling consistent processing pipelines across diverse input formats.

\subsubsection{Content List to Markdown Conversion}
The second stage converts the structured content list into Markdown format. The conversion process iterates over the content list elements and applies type-specific rendering rules that map each semantic type to its corresponding Markdown syntax. The core conversion logic is illustrated with pseudocode in Appendix~\ref{appdx:content_list2markdown}.

\section{Evaluation of MinerU-HTML}
To comprehensively evaluate MinerU-HTML, we construct \textbf{WebMainBench}, a new benchmark designed to assess both main content extraction and structured element preservation. We compare MinerU-HTML against established baseline methods on WebMainBench and validate its generalization capabilities on the external WCEB benchmark.

\subsection{WebMainBench}
\label{sec:WebMainBench}

\subsubsection{Dataset Overview}
WebMainBench comprises 7,887 carefully annotated web pages, each containing: (1) the raw HTML document, (2) ground-truth Main-HTML identified by human annotators as a valid DOM subtree, (3) the corresponding Markdown representation, and (4) rich metadata including language, content type, difficulty level, and structural characteristics. This comprehensive annotation enables fine-grained performance analysis across diverse web content types. An example data entry is shown in Appendix Figure~\ref{fig:benchmark_example}.

\subsubsection{Data Collection and Sampling}
We employ a hybrid sampling strategy to ensure broad coverage of the web ecosystem. We randomly sample 90\% of pages from Common Crawl to capture the long-tail distribution of web content, while the remaining 10\% are drawn from high-traffic websites (Chinaz Alexa\footnote{https://malexa.chinaz.com/}) to include popular, professionally designed pages.

\subsubsection{WebMainBench-Structured: A Subset for Structured Element Extraction}
To rigorously evaluate structured element preservation during main content extraction, we curate \textbf{WebMainBench-Structured}, a focused subset of 545 pages selected from WebMainBench that are rich in mathematical formulas, code blocks, and tables—elements critical for technical and educational content but often corrupted by traditional extraction methods. In addition to standard annotations, each page includes a \texttt{groundtruth\_content} field containing ground-truth Markdown text with all structured elements faithfully preserved by human annotators, enabling precise element-level evaluation.

Table~\ref{tab:elements_distribution} presents the element distribution: 257 pages (47.2\%) contain formulas (predominantly inline, 198 instances), 127 pages (23.3\%) contain code blocks (predominantly interline, 73 instances), and 179 pages (32.8\%) contain tables (predominantly data tables, 151 instances). This distribution reflects realistic usage patterns in technical web content and provides sufficient samples for robust evaluation.

\subsubsection{Annotation Principles}
Defining "main content" is inherently ambiguous, as it varies across different page types and use cases. We establish clear annotation guidelines based on two core principles:

\paragraph{Contextual Integrity.} We include content that is integral to understanding the primary article or page purpose. For example, abstracts, author information, and reference sections are included for academic papers, while related article sidebars and advertisements are excluded.

\paragraph{Human-Generated Content.} We prioritize substantive, human-authored material such as article bodies, user comments, and discussion threads, while excluding auto-generated metadata like timestamps, view counts, and algorithmic recommendations.

Human annotators apply these principles using a custom-built annotation tool (see Appendix Figure~\ref{fig:snapshot}), ensuring consistent and high-quality ground truth. Additional details on the annotation process are provided in Appendix~\ref{appdx:benchmark_construction}.

\subsection{Evaluation Metrics and Baselines}

\subsubsection{Evaluation Metrics}
\paragraph{Main Content Extraction.} We adopt ROUGE-N F1 (N=5) as our primary metric for evaluating overall main content extraction quality. ROUGE-N measures n-gram overlap between predicted and ground-truth content, providing a robust assessment of semantic preservation. We employ the \texttt{jieba} tokenizer for consistent text segmentation. We choose ROUGE-N over ROUGE-L because the Longest Common Subsequence algorithm exhibits prohibitive \(O(n\cdot m)\) computational complexity on the long documents prevalent in our benchmark (average length: 2,304 tokens).

\paragraph{Structured Element Extraction.} To evaluate the fidelity of structured element preservation, we employ specialized metrics for each element type. For \textbf{code blocks} and \textbf{mathematical formulas}, we use normalized edit distance (Levenshtein distance):
\[
\text{EditSim}(s_1, s_2) = 1 - \frac{\text{EditDist}(s_1, s_2)}{\max(|s_1|, |s_2|)}
\]
where \(s_1\) and \(s_2\) are the predicted and ground-truth element strings, and \(|s_1|\) and \(|s_2|\) represent corresponding string length. This metric captures character-level accuracy while normalizing for length variations. For \textbf{tables}, we compute the TEDS (Tree-Edit Distance based Similarity) score~\citep{zhong2020image}, which measures structural similarity by computing tree edit distance on the HTML DOM representation:
\[
\text{TEDS}(T_a, T_b) = 1 - \frac{\text{EditDist}(T_a, T_b)}{\max(|T_a|, |T_b|)}
\]
where \(T_a\) and \(T_b\) are DOM trees of the predicted and ground-truth tables. TEDS is particularly effective for evaluating table structure preservation, as it penalizes row/column misalignment and cell boundary errors.

\subsubsection{Baseline Methods}
We compare MinerU-HTML against two widely adopted extraction systems: \texttt{Trafilatura} \citep{barbaresi2021Trafilatura}, a heuristic-based tool combining DOM analysis and text density heuristics used in RefinedWeb~\citep{refinedweb} and FineWeb~\citep{fineweb}, and \texttt{Resiliparse} \citep{Resiliparse}, a rule-based system optimized for large-scale web archiving used in DCLM \citep{DCLM2024} and Dolma~\citep{dolma}.

\subsection{Results on WebMainBench}

\subsubsection{Main Content Extraction}
Table~\ref{tab:main_results} presents the performance comparison across different content types and difficulty levels. MinerU-HTML achieves an overall ROUGE-N F1 score of 0.8182, establishing state-of-the-art performance and significantly outperforming the best baseline method, Trafilatura (0.6358, Html+MD mode). 

MinerU-HTML demonstrates particularly strong performance on challenging content types where traditional heuristic methods struggle. On pages containing tables, MinerU-HTML achieves 0.7693 compared to Trafilatura's 0.5505. For mathematical equations, MinerU-HTML scores 0.8889 versus 0.7327. Most notably, on conversational content (forums, Q\&A pages), MinerU-HTML achieves 0.7671 compared to just 0.5750 for the best baseline, highlighting the advantage of our semantic, model-based approach over rigid heuristics.

\begin{table}[t!]
  \centering
  \caption{Mean ROUGE-N F1 on WebMainBench with different tracks}
  \resizebox{\textwidth}{!}{%
  \begin{tabular}{llcccccccc}
  \toprule
  name               &   all    &   simple    &   mid    &   hard      &    table    &   code    &   equation     &   conversational    \\
  \midrule
  Resiliparse          &   0.6233 &      0.7099 &   0.6283 &    0.5304  &     0.5473 &         0.6474 &             0.7829 &          0.5346 \\
  Trafilatura           &   0.6358 &      0.7277 &   0.6391 &    0.5396  &     0.5505 &         0.6006 &             0.7327 &          0.5750 \\
  \rowcolor{lightgreen}
  \textbf{MinerU-HTML}            &   \textbf{0.8182} &      \textbf{0.8837} &   \textbf{0.8178} &    \textbf{0.7536}  &     \textbf{0.7693} &         \textbf{0.8368} &             \textbf{0.8889} &          \textbf{0.7671} \\
  \bottomrule
  \end{tabular}
  }
  \label{tab:main_results}
\end{table}

\subsubsection{Efficiency Analysis}
Our pre-processing pipeline achieves substantial token reduction, which is critical for efficient model inference. Compared to a naive generative baseline that would directly process raw HTML (average: 44,706 input tokens, 2,304 output tokens), MinerU-HTML's pre-processing reduces input to 5,735 tokens (12.83\%) and output to 383 tokens (16.64\%). This dramatic reduction makes language model-based extraction computationally feasible and enables deployment of compact models at web scale.

\subsubsection{Structured Element Extraction}
Table~\ref{tab:structured_results} presents a comprehensive evaluation of structured element preservation on WebMainBench-Structured. MinerU-HTML demonstrates exceptional performance across all element types, significantly outperforming both baseline methods.

\begin{figure}[H]
  \centering
  \includegraphics[width=0.9\textwidth]{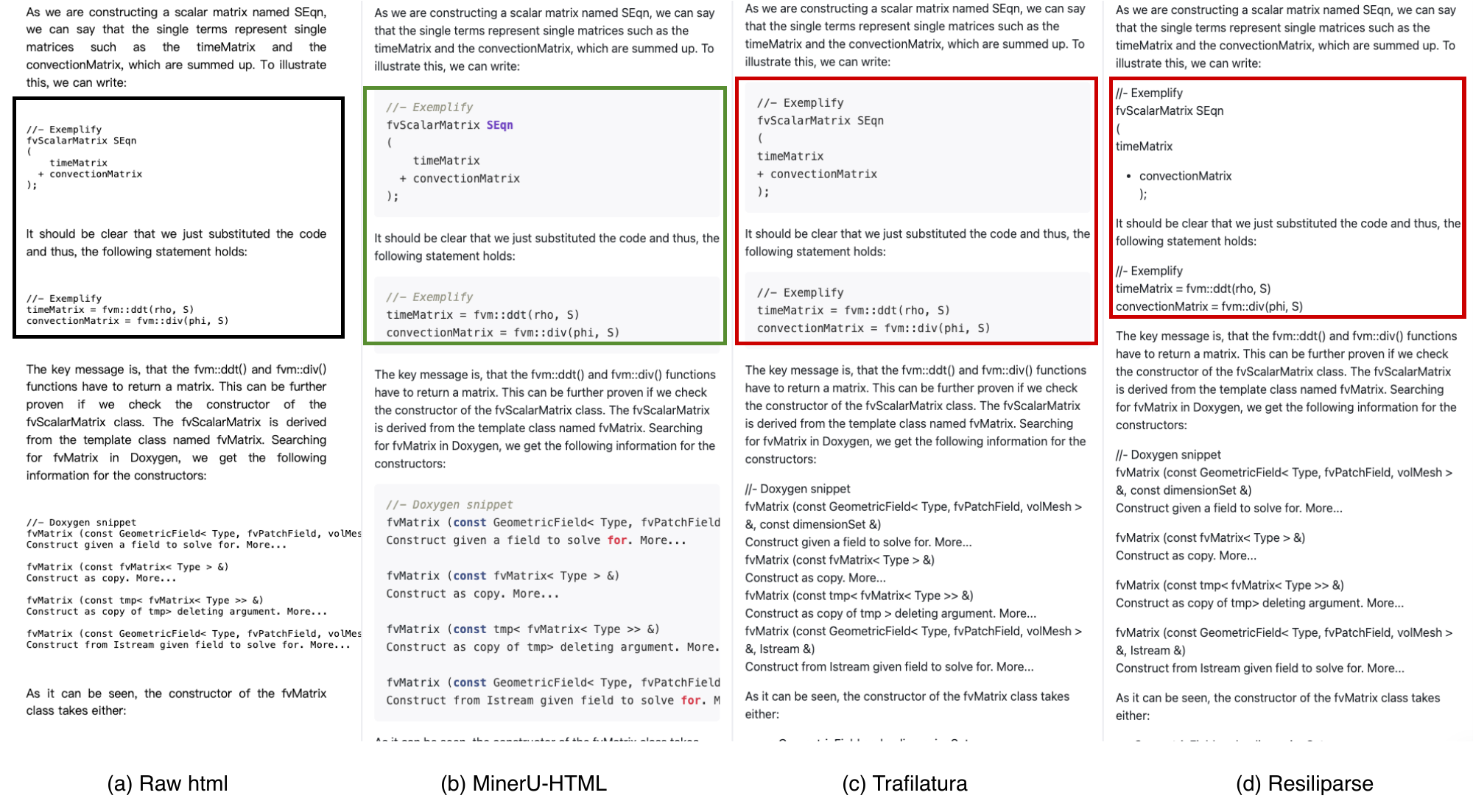}
  \caption{\textbf{Code block extraction comparison.} MinerU-HTML preserves proper indentation and code structure, while Trafilatura strips leading whitespace (converting formatted code into unstructured text) and Resiliparse fails to maintain code block delimiters.}
  \label{fig:case_study_code1}
\end{figure}

\begin{figure}[H]
  \centering
  \includegraphics[width=0.9\textwidth]{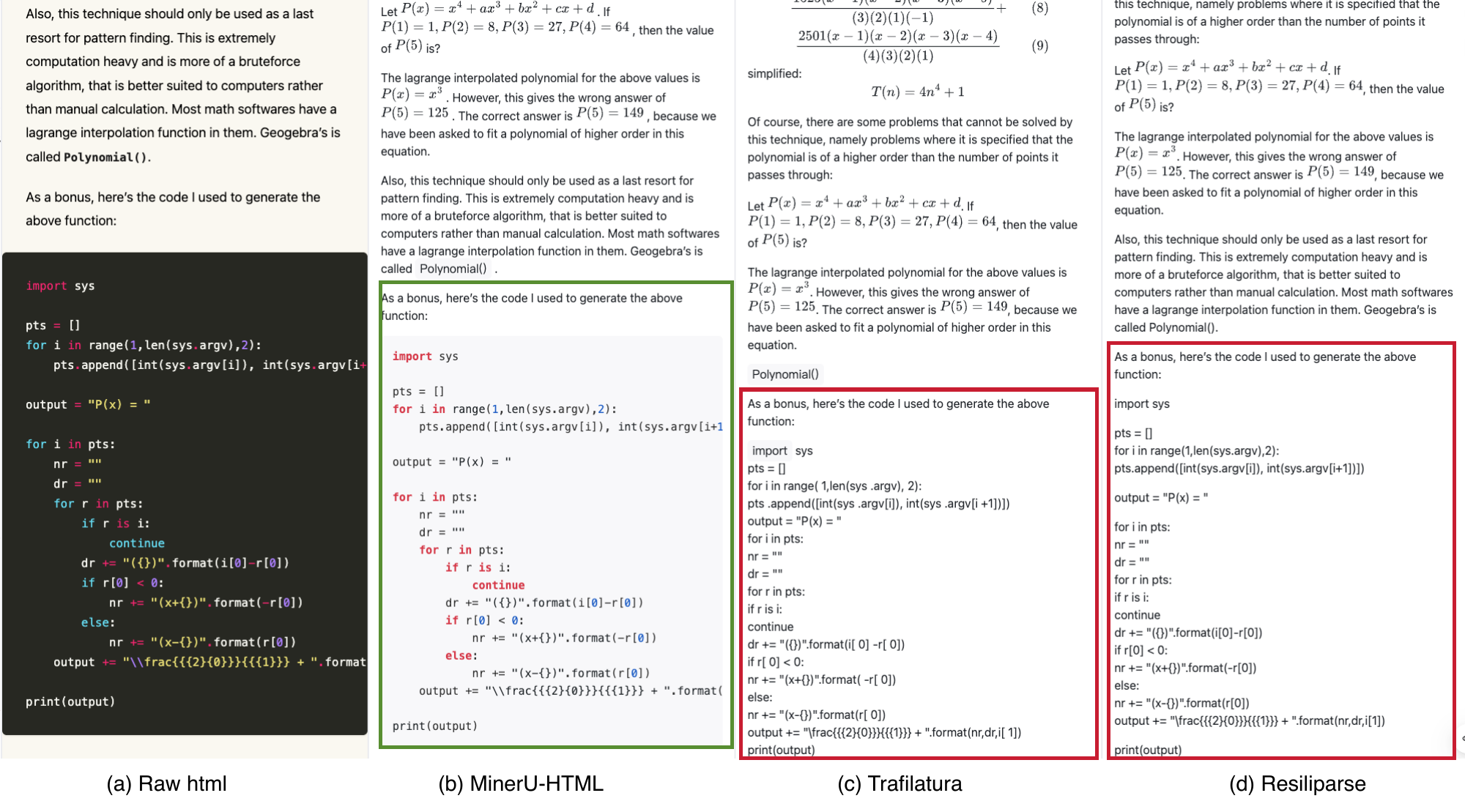}
  \caption{\textbf{Code block extraction comparison.} Both Trafilatura and Resiliparse fail to preserve Markdown code block formatting (\texttt{```} delimiters), rendering code indistinguishable from surrounding text, while MinerU-HTML maintains complete structural integrity.}
  \label{fig:case_study_code2}
\end{figure}

\paragraph{Code Block Preservation.} MinerU-HTML achieves an edit similarity score of 0.9093 for code blocks, substantially exceeding Trafilatura (0.1305) and Resiliparse (0.0641). This dramatic improvement indicates that MinerU-HTML successfully preserves code syntax, indentation, and special characters—elements that are often corrupted by text-density-based heuristics. Qualitative analysis reveals systematic failure patterns in baseline methods: Trafilatura frequently strips indentation, converting properly formatted code into unstructured text (Figure~\ref{fig:case_study_code1}), while Resiliparse often loses Markdown code block delimiters (the \texttt{```} markers), rendering code indistinguishable from surrounding text. Figure~\ref{fig:case_study_code2} demonstrates a representative case where both baselines fail to preserve code block structure, whereas MinerU-HTML maintains complete formatting integrity.

\begin{figure}[H]
  \centering
  \includegraphics[width=0.9\textwidth]{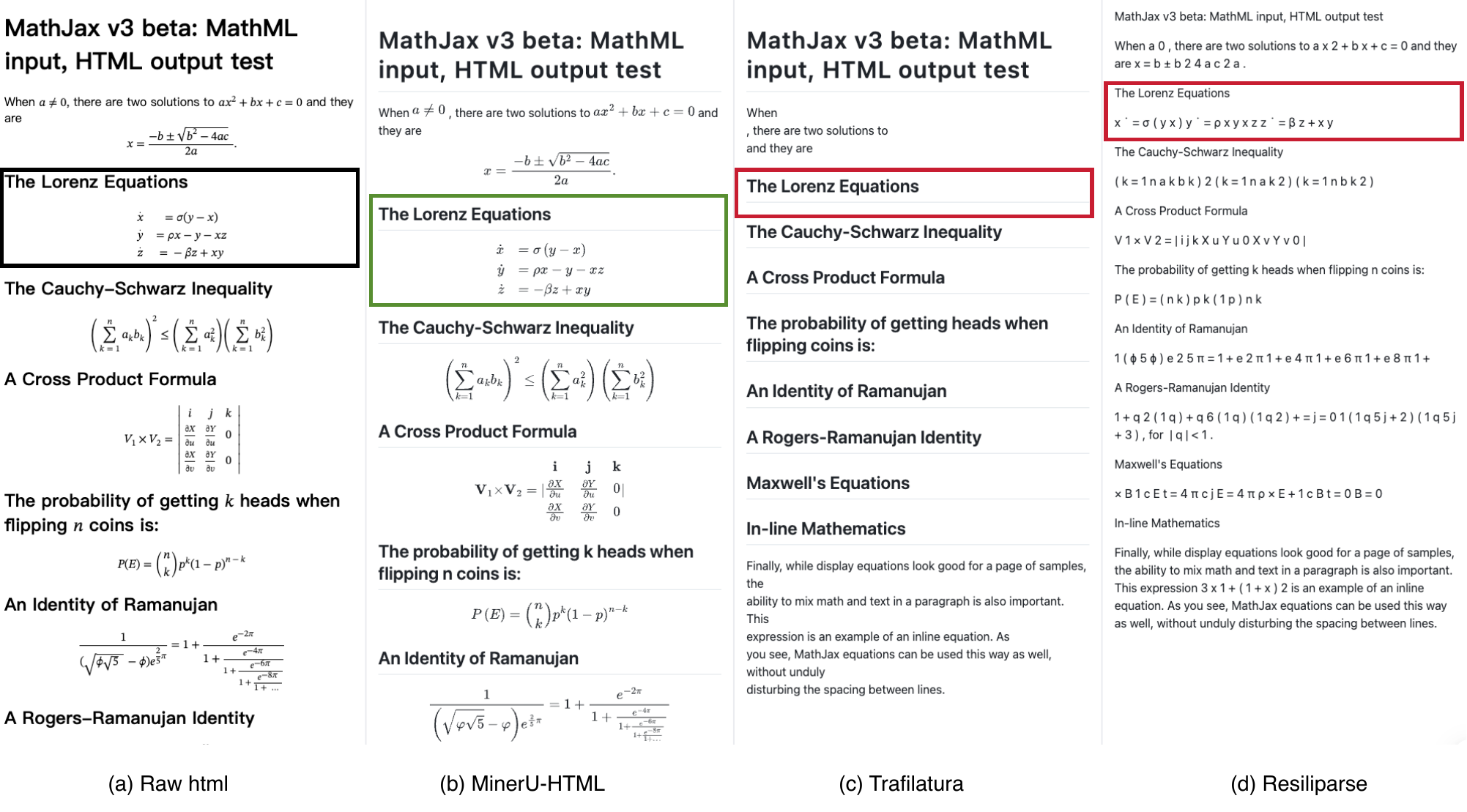}
  \caption{\textbf{Mathematical formula extraction comparison.} Trafilatura fails to recognize and extract the mathematical formulas entirely, discarding them during content extraction. Resiliparse extracts partial content but loses critical LaTeX delimiters and formatting markers. MinerU-HTML correctly preserves the complete formula structure.}
  \label{fig:case_study_math1}
\end{figure}

\begin{figure}[H]
  \centering
  \includegraphics[width=0.9\textwidth]{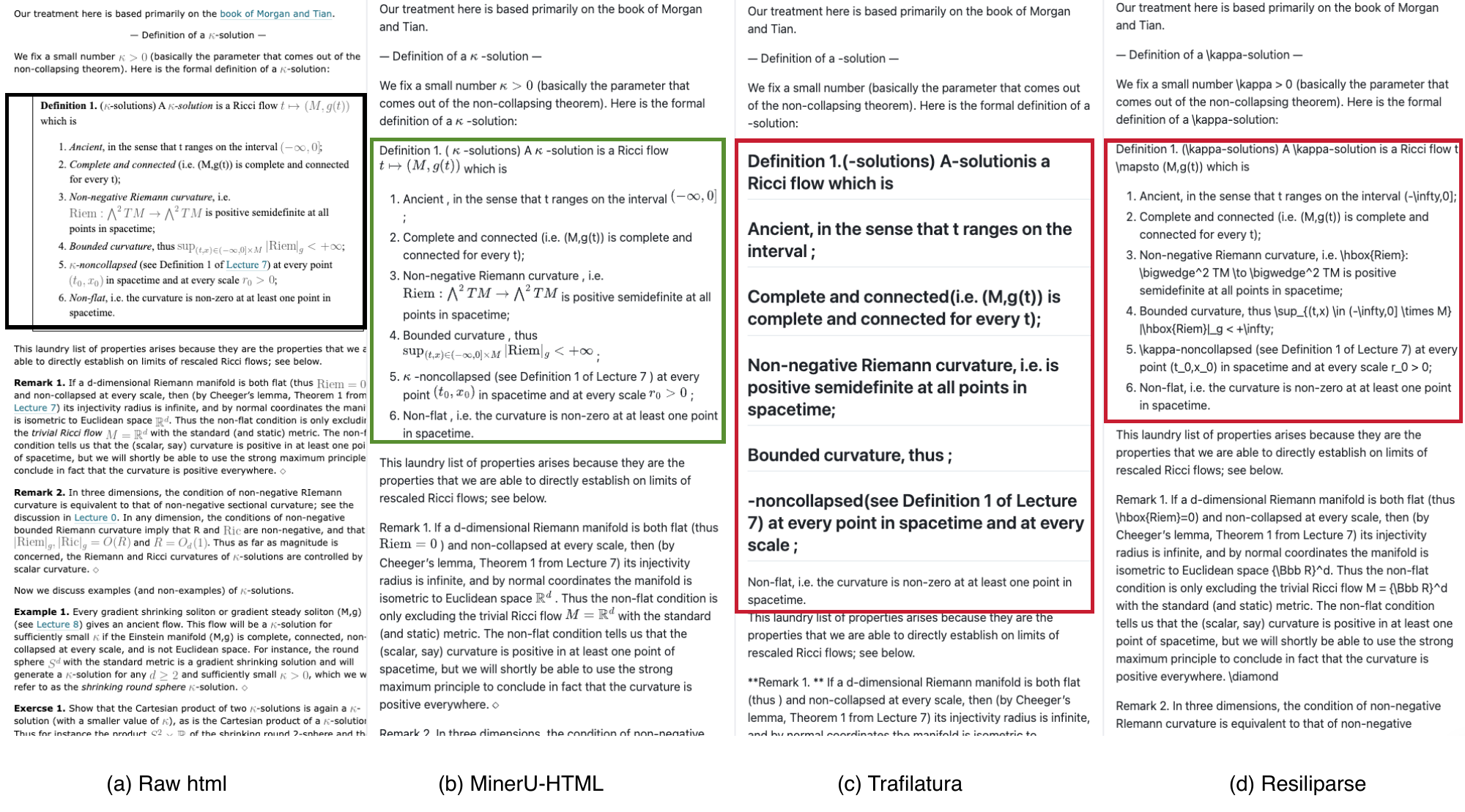}
  \caption{\textbf{Mathematical formula extraction comparison.} Trafilatura misses mathematical notation during extraction, while Resiliparse loses formatting delimiters necessary for proper rendering. MinerU-HTML successfully preserves both inline and display-mode formulas with correct LaTeX syntax.}
  \label{fig:case_study_math2}
\end{figure}

\paragraph{Mathematical Formula Preservation.} For mathematical formulas, MinerU-HTML achieves 0.9399 edit similarity, compared to 0.6107 (Trafilatura) and 0.6778 (Resiliparse). This high score reflects MinerU-HTML's ability to correctly identify and preserve LaTeX expressions, MathML markup, and inline mathematical notation across diverse rendering engines. The baseline methods exhibit two primary failure modes: Trafilatura often fails to recognize mathematical content entirely, discarding formulas during extraction (Figures~\ref{fig:case_study_math1} and~\ref{fig:case_study_math2}), while Resiliparse extracts the textual components but loses critical formatting delimiters and structural markers, rendering the mathematical notation unparseable by downstream processors.

\begin{figure}[H]
  \centering
  \includegraphics[width=0.9\textwidth]{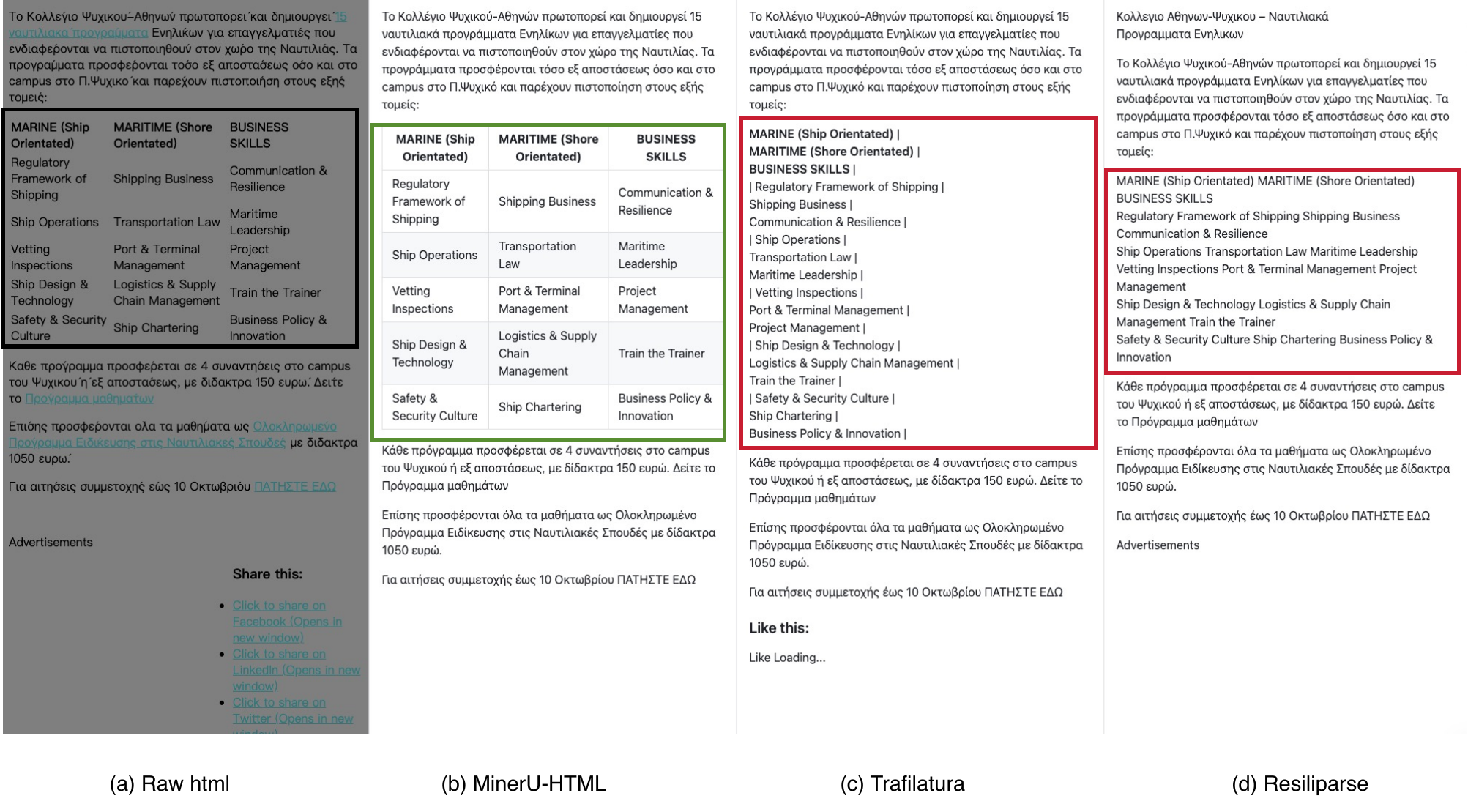}
  \caption{\textbf{Table structure extraction comparison.} Trafilatura and Resiliparse fail to recognize table boundaries and cell relationships, linearizing tabular data into unstructured text. MinerU-HTML preserves the table structure, maintaining row-column organization and cell alignment.}
  \label{fig:case_study_table1}
\end{figure}

\begin{figure}[H]
  \centering
  \includegraphics[width=0.9\textwidth]{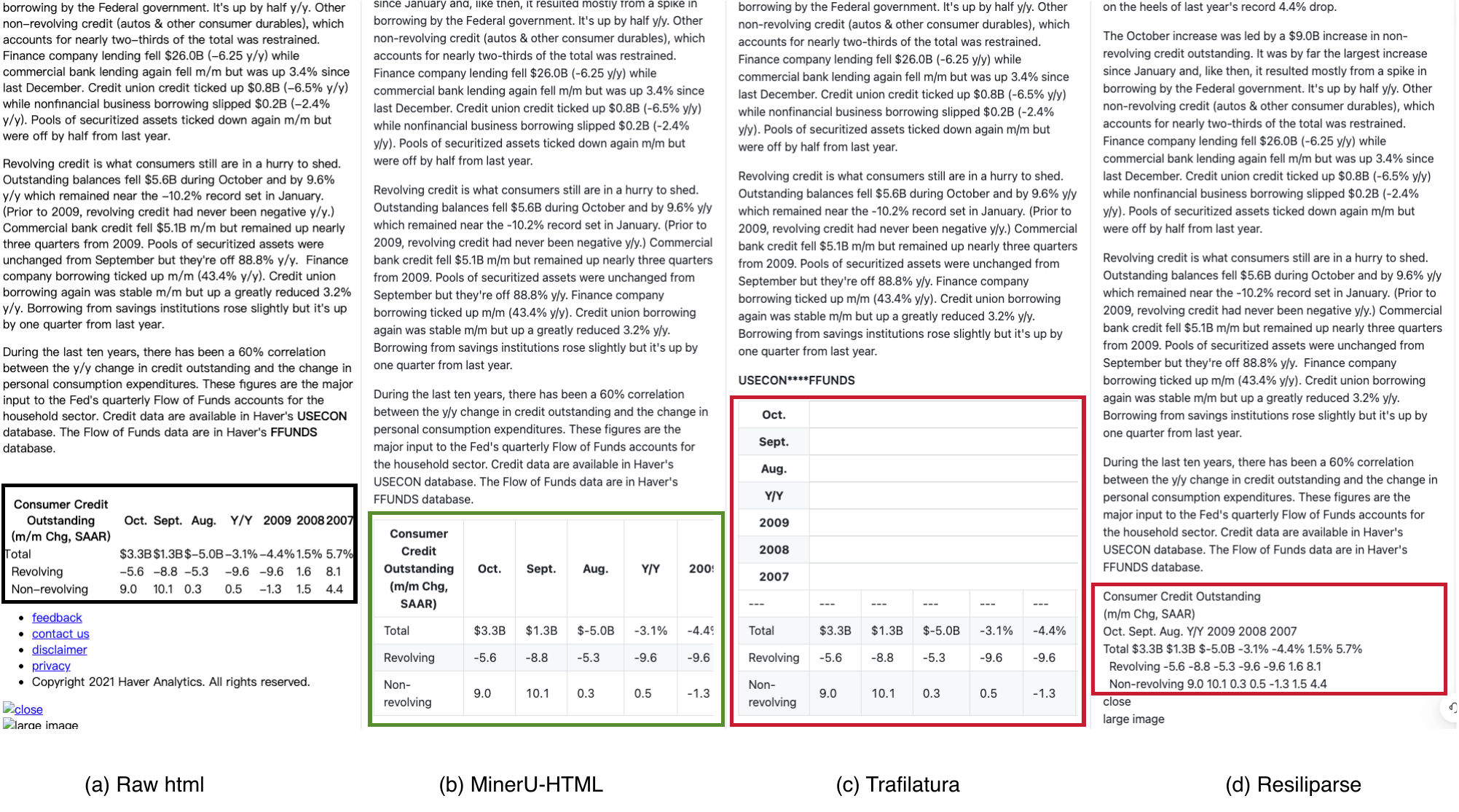}
  \caption{\textbf{Table structure extraction comparison.} Both baseline methods collapse the table structure, losing critical semantic relationships between data cells. MinerU-HTML maintains structural fidelity, preserving headers, rows, and cell boundaries necessary for downstream table understanding.}
  \label{fig:case_study_table2}
\end{figure}

\paragraph{Table Structure Preservation.} For table extraction, we evaluate structural preservation using the TEDS metric. MinerU-HTML achieves 0.7388, significantly outperforming Trafilatura (0.3405) and Resiliparse (0.0227). While this score is lower compared to code and formula extraction, this reflects the inherent complexity of table structure preservation rather than a fundamental limitation of our approach. Complex table layouts with merged cells, nested structures, and inconsistent markup remain challenging even for advanced extraction methods. Nevertheless, MinerU-HTML's 2.2$\times$ improvement over Trafilatura and 32.5$\times$ improvement over Resiliparse demonstrates substantial progress. As illustrated in Figures~\ref{fig:case_study_table1} and~\ref{fig:case_study_table2}, baseline methods consistently fail to recognize table boundaries and cell relationships, often linearizing tabular data into unstructured paragraphs, while MinerU-HTML preserves row-column structure and cell alignment.

\paragraph{Implications.} These results validate that MinerU-HTML's Main-HTML extraction pipeline, combining semantic understanding with structural awareness, effectively preserves complex content structures that are crucial for downstream applications such as language model pretraining, technical documentation archiving, and knowledge base construction. The preservation of structured elements is particularly important for training models on technical and scientific content, where code, formulas, and tables convey information that cannot be adequately represented in plain text.

\begin{table}[t]
  \centering
  \caption{Structured Element Extraction Performance on WebMainBench-Structured. Edit similarity scores (higher is better) for code blocks and formulas. Table TEDS measures structural preservation for tables.}
  \label{tab:structured_results}
  \begin{tabular}{llll}
  \toprule
  \textbf{Extractor} & \textbf{Code Edit} & \textbf{Formula Edit} & \textbf{Table TEDS} \\
  \midrule
  Trafilatura        & 0.1305             & 0.6107                & 0.3405              \\
  Resiliparse        & 0.0641             & 0.6778                & 0.0227              \\
  \rowcolor{lightgreen}
  \textbf{MinerU-HTML}   & \textbf{0.9093}    & \textbf{0.9399}       & \textbf{0.7388}     \\
  \bottomrule
  \end{tabular}
\end{table}

\subsection{Generalization to WCEB}

To assess MinerU-HTML's generalization capabilities beyond our benchmark, we evaluate on the Web Content Extraction Benchmark (WCEB)~\citep{survey}, a unified collection of nine established datasets. WCEB provides cleaned and standardized ground truths, addressing inconsistencies prevalent in legacy datasets such as encoding errors and script injection artifacts. A detailed description of the benchmark can be found in Table~\ref{tab:datasets}. Since WCEB uses plain-text ground truths, we convert the extracted MAIN-HTML by MinerU-HTML to plain-text using the \texttt{html-text} library\footnote{https://pypi.org/project/html-text/}.

\begin{table}[t]
    \centering
    \caption{Results on WCEB}
    \label{tab:WCEB_results}
    \begin{tabular}{lcccc}
        \toprule
        name & all & simple & mid & hard \\
        \midrule
        Trafilatura & 0.7833 & 0.8122 & 0.7785 & 0.7609 \\
        Resiliparse & 0.7225 & 0.7697 & 0.7052 & 0.6985 \\
        \rowcolor{lightgreen}
        \textbf{MinerU-HTML} & \textbf{0.8002} & \textbf{0.8293} & \textbf{0.8005} & \textbf{0.7707} \\
        \bottomrule
    \end{tabular}
\end{table}

Table~\ref{tab:WCEB_results} shows that MinerU-HTML maintains strong performance on WCEB, achieving an overall score of 0.8002 and outperforming the strongest baseline, Trafilatura (0.7833). 
The strong performance across both WebMainBench and WCEB validates MinerU-HTML's robustness and generalization capabilities across diverse web content.

\section{Dataset Construction and Pretraining Experiments}
\label{sec:pretraining}
To validate the effectiveness of MinerU-HTML for downstream language model training, we construct AICC from Common Crawl and conduct comprehensive pretraining experiments comparing it against strong baselines. This section describes our dataset construction pipeline, experimental setup, and downstream evaluation results.

\subsection{Dataset Construction}
\label{subsec:dataset_construction}

\subsubsection{Extraction and Initial Processing}
We apply MinerU-HTML to extract content from two Common Crawl snapshots (CC-2023-06 and CC-2023-14)\footnote{To enable direct comparison with existing datasets such as RefinedWeb and FineWeb, we use CC-2023-06 and CC-2023-14 snapshots in this paper. The released version of AICC is based on more recent snapshots (CC-2025-08 and CC-2025-13) to provide the community with the most up-to-date data.}, converting raw HTML WARC archives into Markdown format. This extraction process yields \textbf{AICC}, a large-scale multilingual web corpus optimized for language model training. For rigorous baseline comparison, we also construct \textbf{TfCC} by applying Trafilatura—a widely-adopted extraction tool—to the identical two snapshots under the same processing pipeline.

\begin{figure}[t]
  \centering
  \caption{\textbf{Length ratio distribution between AICC and TfCC documents}. Positive values indicate AICC extracts more content.}
  \label{fig:length_ratio_distribution}
  \includegraphics[width=0.8\textwidth]{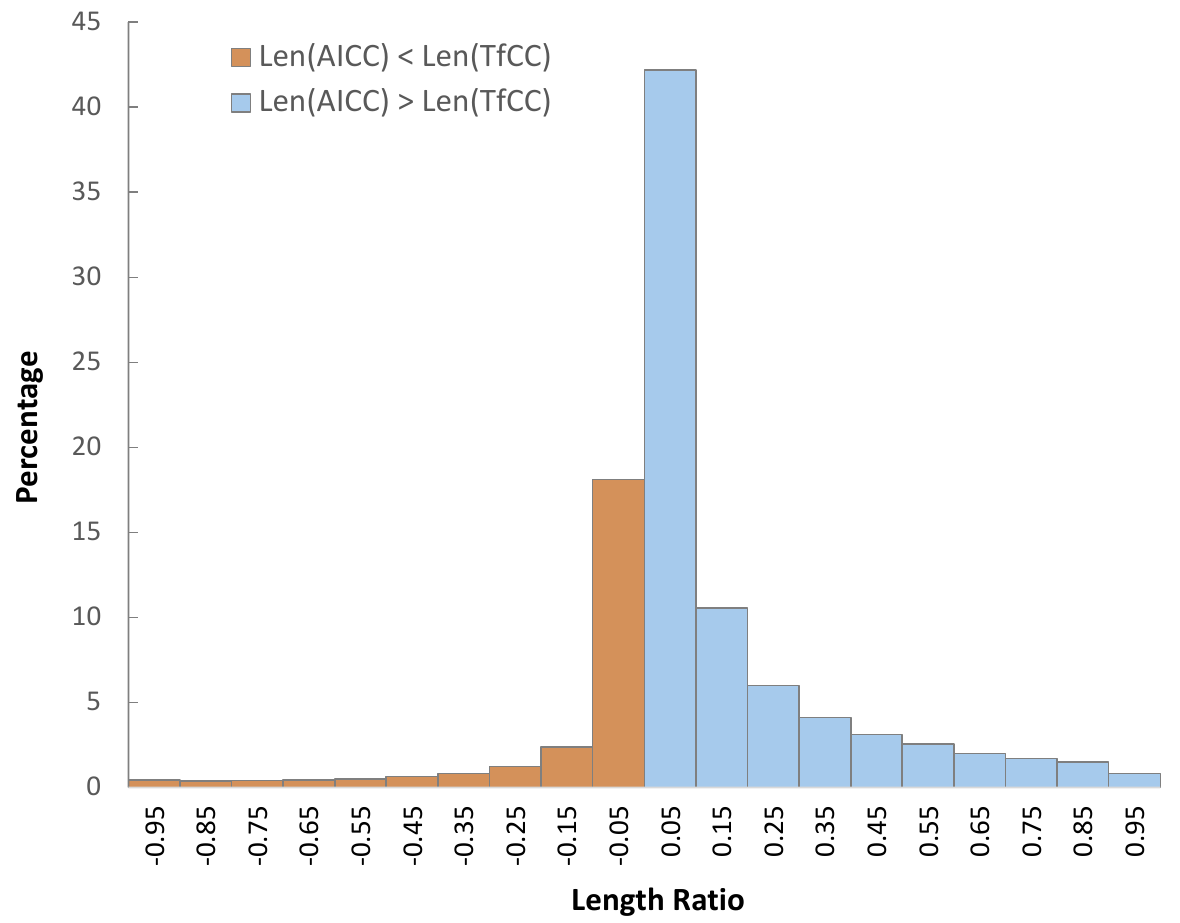}
\end{figure}

To quantify extraction quality differences, we sample 800k document pairs and compute length ratios:
\[
\text{Length Ratio} = \frac{\text{Len}_{\text{AICC}} - \text{Len}_{\text{TfCC}}}{\max(\text{Len}_{\text{AICC}}, \text{Len}_{\text{TfCC}})}
\]
where \(\text{Len}_{\text{AICC}}\) and \(\text{Len}_{\text{TfCC}}\) represent the length of extracted content from AICC and TfCC documents, respectively. The distribution (Figure~\ref{fig:length_ratio_distribution}) reveals that AICC documents are 1.16$\times$ longer on average, indicating that MinerU-HTML preserves more content.

\subsubsection{Extraction Quality Analysis}
We evaluate how extraction length relates to perceived quality using an LLM-as-a-judge protocol \citep{dingo} with DeepSeek-Chat-V3, see the prompt in Figure~\ref{fig:rating_prompt}. We draw a stratified sample of 10{,}000 AICC vs.~TfCC document pairs uniformly across 20 length-ratio bins (from $-1.0$ to $1.0$; $\sim$500 per bin) to ensure balanced coverage across the full spectrum of length differences.

\begin{figure}[t]
  \centering
  \includegraphics[width=0.9\textwidth]{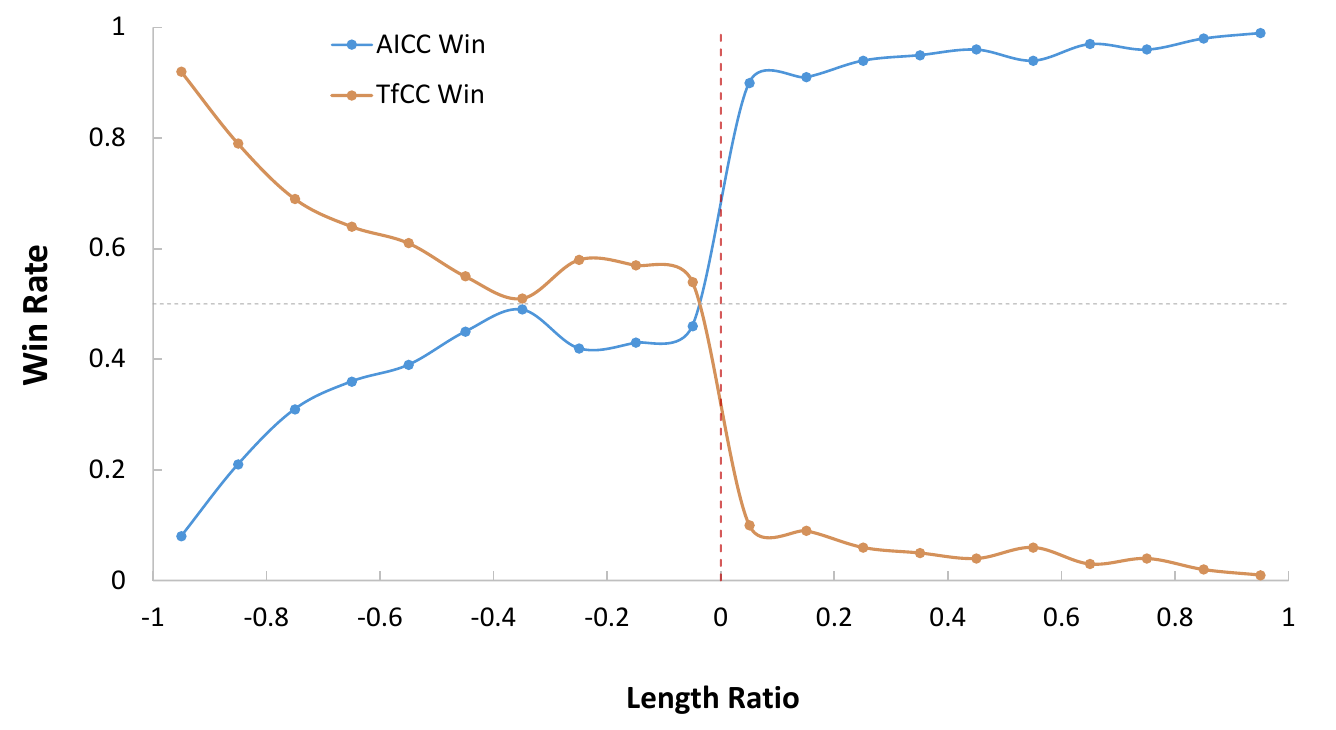}
  \caption{\textbf{Extraction quality vs.~length ratio.} Pairwise win rates (AICC vs.~TfCC) judged by DeepSeek-Chat-V3 on 10{,}000 stratified samples. A sharp crossover at ratio = 0 (red dashed line) reveals that when AICC extracts more content (positive ratios), it is preferred in 75–98\% of comparisons; when it extracts less (negative ratios), TfCC is preferred in 51–92\% of cases.}
  \label{fig:quality_vs_length_ratio}
\end{figure}

Figure~\ref{fig:quality_vs_length_ratio} shows a sharp crossover at length ratio $0$. When AICC extracts less than TfCC (negative ratios), TfCC is preferred with win rates of 51–92\%. When AICC extracts more (positive ratios), the preference reverses: AICC wins 75–98\%, with preference increasing monotonically with the ratio.

Weighting the bin-wise win rates by the population distribution in Figure~\ref{fig:length_ratio_distribution}, AICC achieves a 72.0\% overall win rate. This indicates that the additional content preserved by MinerU-HTML is overwhelmingly judged as valuable main content rather than noise, reflecting both higher recall and maintained precision relative to heuristic extraction.

\begin{figure}[H]
  \centering
  \includegraphics[width=0.9\textwidth]{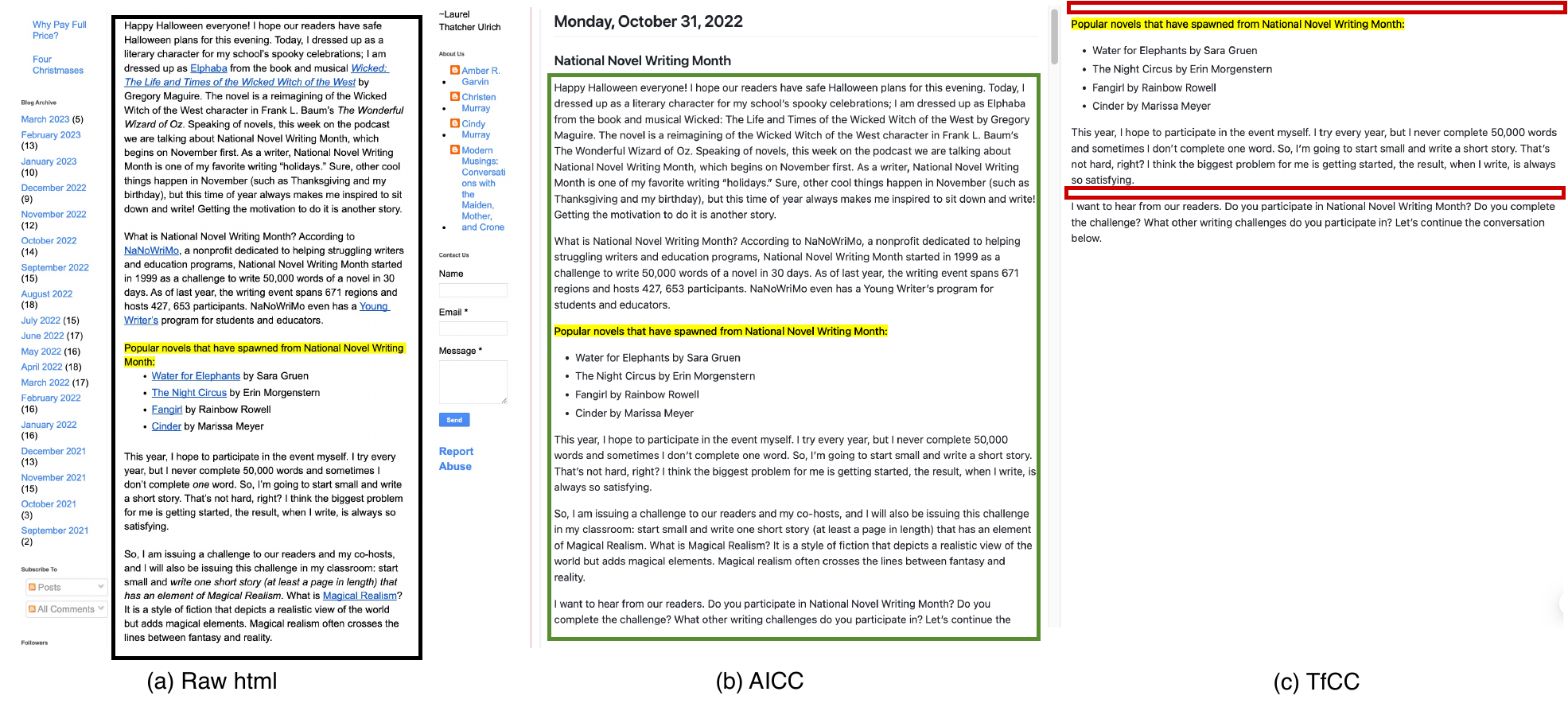}
  \caption{AICC longer than TfCC. The AICC document is preferred over the TfCC document.}
  \label{fig:AICC_win_Tshorter1}
\end{figure}

\begin{figure}[H]
  \centering
  \includegraphics[width=0.9\textwidth]{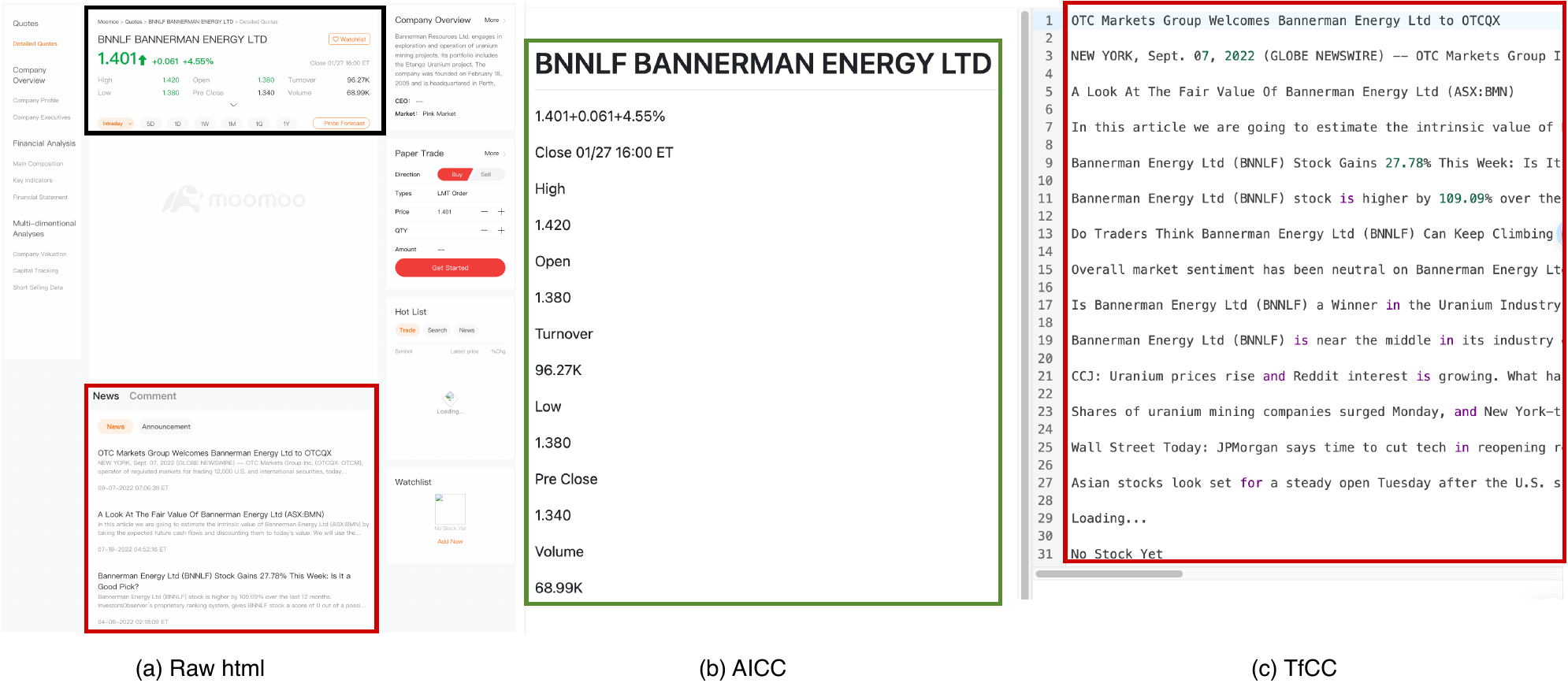}
  \caption{TfCC longer than AICC. The AICC document is preferred over the TfCC document.}
  \label{fig:AICC_win_Tlonger1}
\end{figure}

\begin{figure}[H]
  \centering
  \includegraphics[width=0.9\textwidth]{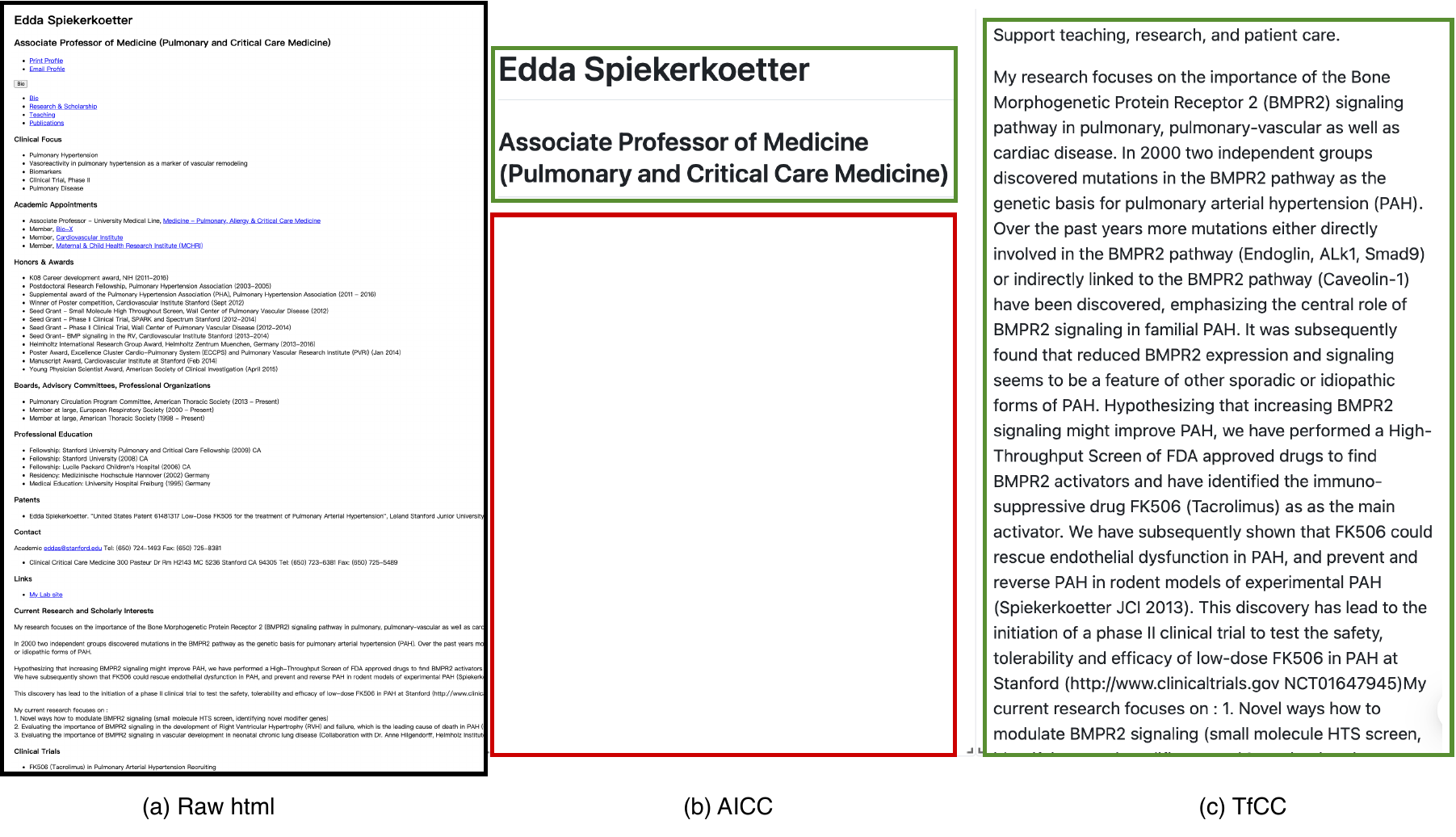}
  \caption{TfCC longer than AICC. The TfCC document is preferred over the AICC document.}
  \label{fig:AICC_lose_Tlonger1}
\end{figure}

\begin{figure}[H]
  \centering
  \includegraphics[width=0.9\textwidth]{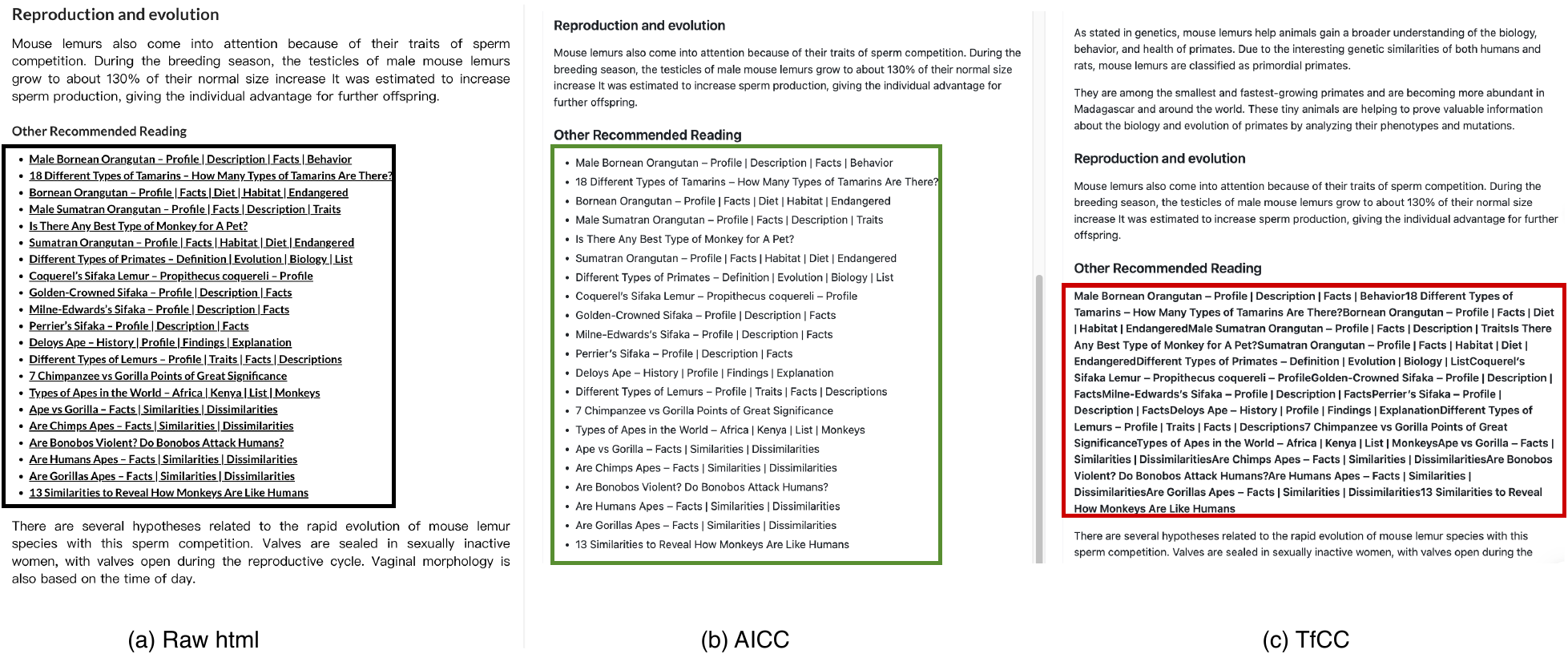}
  \caption{Similar length. The AICC document is preferred over the TfCC document.}
  \label{fig:AICC_win_similar_length1}
\end{figure}

\begin{figure}[H]
  \centering
  \includegraphics[width=0.9\textwidth]{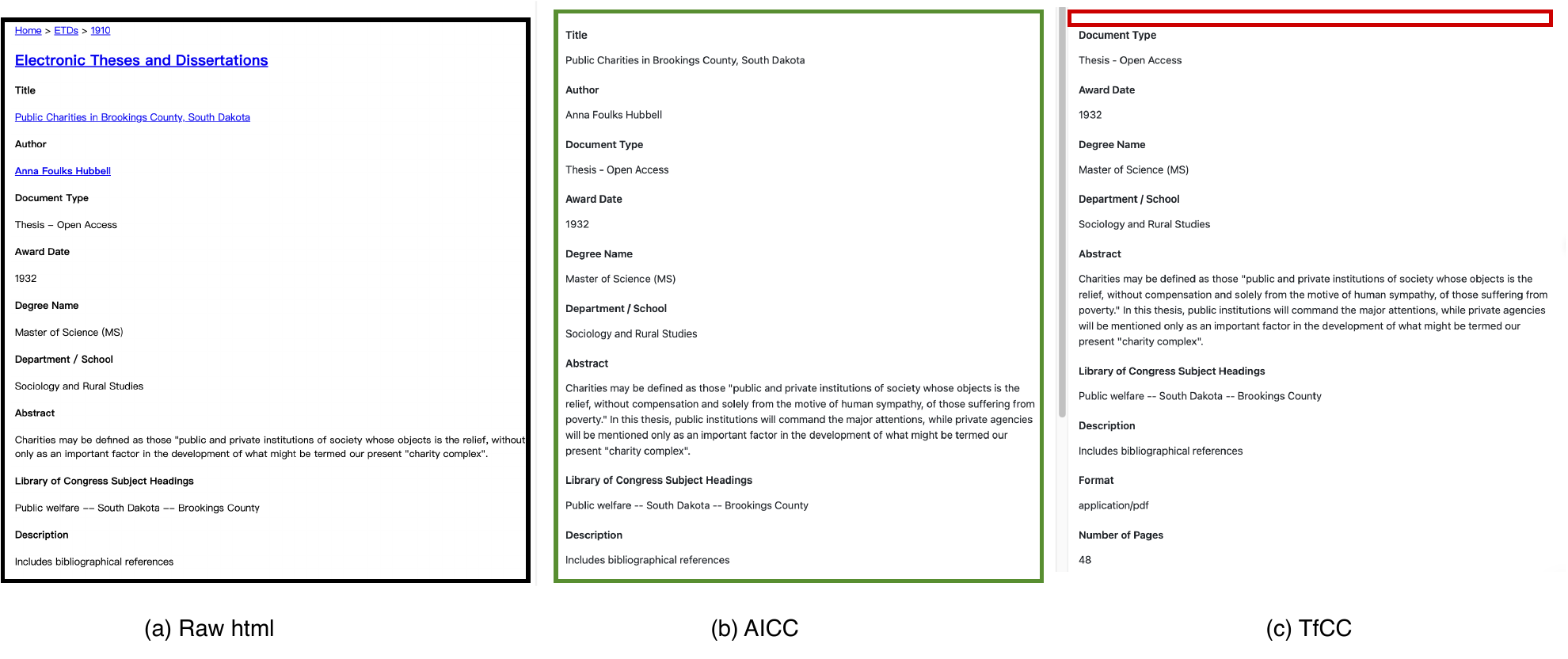}
  \caption{Similar length. The AICC document is preferred over the TfCC document.}
  \label{fig:AICC_win_similar_length2}
\end{figure}

To understand failure modes, we manually examine cases from the extreme and near-zero bins, with examples provided in Appendix~\ref{appdx:AICC-TfCC_comparison}:
\begin{enumerate}[leftmargin=*,topsep=0pt,itemsep=2pt]
  \item \textbf{[0.9, 1.0]: AICC longer than TfCC}: Most AICC documents are preferred. For example, Figure~\ref{fig:AICC_win_Tshorter1} shows a case where Trafilatura drops much of the main body content.
  \item \textbf{[-0.1, 0.1]: Similar length}: Even at similar lengths, AICC can be superior. In one case (Figure~\ref{fig:AICC_win_similar_length1}), Trafilatura mishandles list items; in another (Figure~\ref{fig:AICC_win_similar_length2}), it misses important elements such as the title and author on a dissertation page.
  \item \textbf{[-1.0, -0.9]: TfCC longer than AICC}: TfCC often wins because AICC misses portions of the main content (see Figure~\ref{fig:AICC_lose_Tlonger1}). However, there are also instances where Trafilatura retains substantial boilerplate (e.g., advertisements) that MinerU-HTML correctly excludes (see Figure~\ref{fig:AICC_win_Tlonger1}).
\end{enumerate}

While AICC does exhibit failure cases, the overall quality advantage over TfCC remains robust, as evidenced by the 72.0\% win rate and the consistent performance gains observed in downstream pretraining experiments (Section~\ref{subsec:results}).

\subsubsection{Quality Filtering and Deduplication}
To prepare both datasets for pretraining, we apply identical post-processing pipelines following established practices from RefinedWeb~\citep{refinedweb}: (1) exact deduplication using sha256 (2) language identification and filtering using FastText, (3) quality filtering using Gopher heuristic rules~\citep{gopher}, (4) safety filtering with URL and keyword blacklists, and (5) fuzzy deduplication using MinHash with Locality-Sensitive Hashing. This pipeline reduces the corpora to \textbf{372B tokens} (AICC) and \textbf{317B tokens} (TfCC). The identical filtering ensures that any downstream performance differences can be attributed solely to extraction quality rather than filtering strategies.

\subsection{Pretraining Setup}
\label{subsec:setup}

\paragraph{Training.}
The models utilized in this work are 1.5B parameter decoder-only transformers derived from the Qwen3 model family.  
For the standard pretraining configuration, each model is trained from scratch on a selected subset of 62B tokens.
A comprehensive description of the architecture and training setup is available in Appendix \ref{appdx:pretrain_detail}.

\paragraph{Evaluation Benchmarks.}
We evaluate pretrained models through in-context learning using the \textit{lm-evaluation-harness} framework \citep{eval-harness}, reporting accuracy as the primary metric. The evaluation suite spans three task categories with 13 benchmarks: (1) \textbf{General Knowledge}—ARC \citep{arc}, BoolQ \citep{boolq}, CommonsenseQA \citep{commonsenseqa}, MMLU \citep{mmlu}, and SciQ \citep{sciq}; (2) \textbf{Reasoning}—COPA \citep{gordon2012semeval}, HellaSwag \citep{hellaswag}, PIQA \citep{bisk2020piqa}, SIQA \citep{siqa}, and WinoGrande \citep{winogrande}; (3) \textbf{Reading Comprehension}—CoQA \citep{reddy2019coqa}, LAMBADA \citep{paperno2016lambada}, and OpenBookQA \citep{openbookqa}. Further evaluation details are provided in Appendix \ref{appdx:evaluation}.

\paragraph{Baseline Datasets.}
We compare AICC against three strong baselines:
\begin{enumerate}[leftmargin=*,topsep=0pt,itemsep=2pt]
\item \textbf{TfCC}: Applies Trafilatura extraction to the same CC snapshots with identical post-processing (Section~\ref{subsec:dataset_construction}). This controlled comparison isolates extraction quality's impact by eliminating confounding factors from filtering strategies.

\item \textbf{RefinedWeb}~\citep{refinedweb}: A widely-adopted web corpus from Technology Innovation Institute using aggressive filtering and deduplication on Common Crawl. RefinedWeb has demonstrated performance comparable to curated corpora like C4 and serves as a strong heuristic-based baseline.

\item \textbf{FineWeb}~\citep{fineweb}: A state-of-the-art 15-trillion token corpus from HuggingFace with optimized filtering and deduplication pipelines. FineWeb represents the current best practice for heuristic-based web corpus construction and outperforms RefinedWeb, C4, and Dolma.
\end{enumerate}

We exclude datasets employing model-based quality filtering (e.g., DCLM-Baseline~\citep{DCLM2024}), as this introduces confounding factors that obscure extraction quality's specific contribution.

\subsection{Downstream Evaluation Results}
\label{subsec:results}
\begin{figure}[t]
  \centering
  \caption{\textbf{Training dynamics across 13 benchmarks for models pretrained on AICC and TfCC.} Average accuracy at 15 checkpoints (4B–63B tokens). AICC consistently maintains superior or competitive performance throughout training.}
  \label{fig:Perf_2methods_dynamic}
  \includegraphics[width=0.8\textwidth]{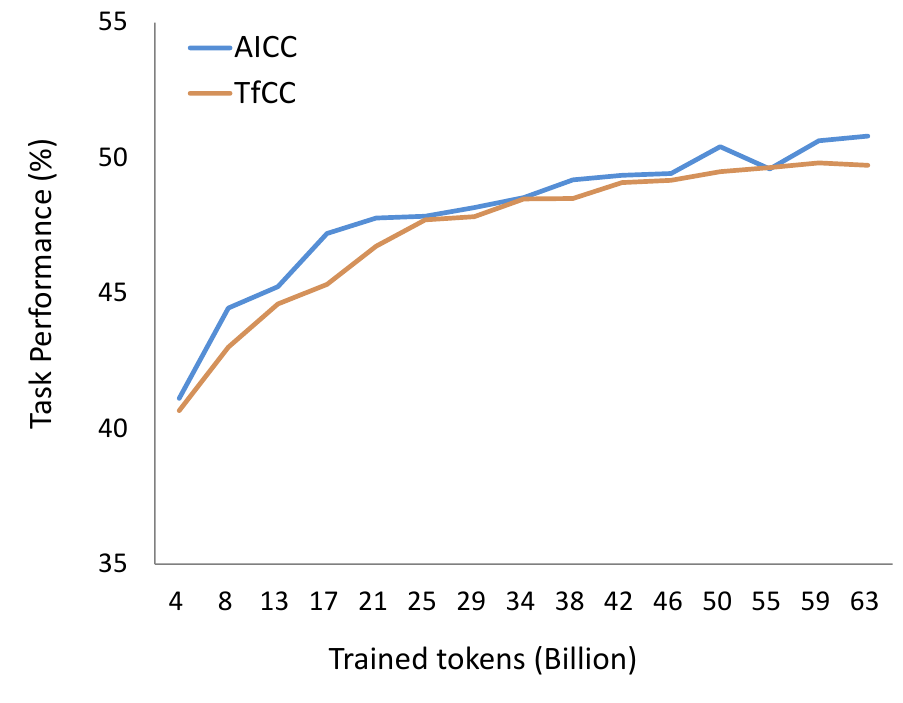}
\end{figure}

\begin{figure}[t]
  \centering
  \caption{\textbf{Task category breakdown comparing AICC, TfCC, RefinedWeb, and FineWeb at 63B tokens.} AICC achieves the best performance on General Knowledge and Reading Comprehension tasks, significantly outperforming FineWeb on Reading Comprehension (+5.69pp). General Knowl. = General Knowledge, Reading Compr. = Reading Comprehension.}
  \label{fig:Perf_4methods_last_ckpt}
  \includegraphics[width=0.8\textwidth]{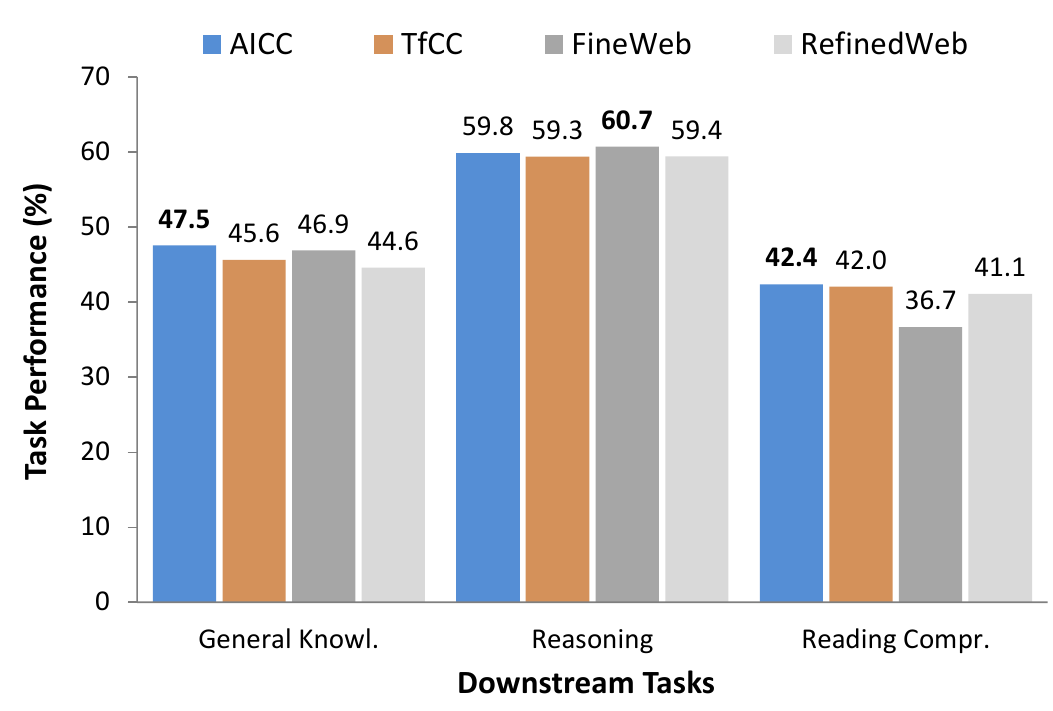}
\end{figure}

We compare models pretrained on AICC (MinerU-HTML-extracted) against TfCC (Trafilatura-extracted), RefinedWeb, and FineWeb across 13 benchmarks. Figure~\ref{fig:Perf_2methods_dynamic} and Table~\ref{tab:Perf_2methods_dynamic} present training dynamics at 15 checkpoints (4B–63B tokens), while Figure~\ref{fig:Perf_4methods_last_ckpt} and Table~\ref{tab:Perf_4methods_last_ckpt} provide detailed task category comparisons between all four corpora at the final checkpoint.

\paragraph{Overall Performance and Training Dynamics.} At the final checkpoint (63B tokens), models trained on AICC achieve 50.82\% average accuracy, outperforming FineWeb (49.61\%), RefinedWeb (49.13\%), and TfCC (49.74\%). The 1.08pp improvement over TfCC provides direct evidence that extraction quality significantly impacts downstream capabilities, since both datasets differ only in extraction method while undergoing identical post-processing. Notably, AICC surpasses even FineWeb—a state-of-the-art corpus with sophisticated quality filtering—demonstrating that semantic-aware extraction can rival aggressive filtering strategies. Figure~\ref{fig:Perf_2methods_dynamic} and Table~\ref{tab:Perf_2methods_dynamic} show this advantage persists throughout training: from the earliest checkpoint (4B tokens) to the final checkpoint, AICC consistently maintains superior or competitive performance, indicating stable quality improvements across the full training trajectory.

\paragraph{Task Category Analysis.} Table~\ref{tab:Perf_4methods_last_ckpt} and Figure~\ref{fig:Perf_4methods_last_ckpt} provide a comprehensive comparison across all four datasets. The AICC vs. TfCC comparison is particularly informative for isolating extraction quality's impact, revealing consistent improvements across task categories: \textbf{General Knowledge} improves by 1.93pp (47.54\% vs. 45.61\%), \textbf{Reasoning} by 0.49pp (59.83\% vs. 59.34\%), and \textbf{Reading Comprehension} by 0.35pp (42.37\% vs. 42.02\%). Notably, AICC achieves the best performance among all methods on General Knowledge tasks, and competitive performance on Reading Comprehension. Compared to the external baselines, AICC outperforms FineWeb on General Knowledge (47.54\% vs. 46.86\%) and Reading Comprehension (42.37\% vs. 36.68\%), while FineWeb shows slightly better performance on Reasoning tasks (60.69\% vs. 59.83\%). These results validate that extraction quality significantly impacts model capabilities, with semantic-aware extraction complementing aggressive filtering strategies.

\section{Conclusion}

This work demonstrates that HTML-to-text extraction is a critical but under-studied component of web corpus construction. Our controlled experiments show that extraction quality alone yields performance gains comparable to sophisticated filtering strategies: models trained on AICC achieve 1.08pp improvement over TfCC (both with identical post-processing) and outperform RefinedWeb and FineWeb across multiple benchmarks. Large-scale validation on our 4.52-trillion token corpus confirms that AICC is preferred over TfCC in 72.0\% of 10,000 document comparisons, with MinerU-HTML preserving 1.16$\times$ more content judged as valuable main content rather than noise.

A key insight is that model-based extraction offers inherent scalability advantages: while heuristic extractors face fundamental limitations where rule updates often conflict, MinerU-HTML's performance can systematically improve with better data and models. This positions extraction quality as a continuously improvable component rather than a fixed constraint. Future work should explore: (1) JavaScript rendering for single-page applications; (2) learned clustering methods for improved template detection; (3) larger-scale pretraining validation (10B+ parameters); (4) multi-modal extraction; and (5) integration with model-based quality filtering.

We release MinerU-HTML, WebMainBench, and AICC to facilitate further research into semantic-aware extraction methods and their impact on pretraining data quality.

\clearpage
\newpage
\bibliography{main}

@article{brown2020language,
  title={Language models are few-shot learners},
  author={Brown, Tom and Mann, Benjamin and Ryder, Nick and Subbiah, Melanie and Kaplan, Jared D and Dhariwal, Prafulla and Neelakantan, Arvind and Shyam, Pranav and Sastry, Girish and Askell, Amanda and others},
  journal={Advances in neural information processing systems},
  volume={33},
  pages={1877--1901},
  year={2020}
}

@inproceedings{zhong2020image,
  title={Image-based table recognition: data, model, and evaluation},
  author={Zhong, Xu and ShafieiBavani, Elaheh and Jimeno Yepes, Antonio},
  booktitle={European conference on computer vision},
  pages={564--580},
  year={2020},
  organization={Springer}
}

@InProceedings{Resiliparse,
  address =             {Berlin Heidelberg New York},
  author =              {Janek Bevendorff and Benno Stein and Matthias Hagen and Martin Potthast},
  booktitle =           {Advances in Information Retrieval. 40th European Conference on IR Research (ECIR 2018)},
  editor =              {Leif Azzopardi and Allan Hanbury and Gabriella Pasi and Benjamin Piwowarski},
  month =               mar,
  publisher =           {Springer},
  series =              {Lecture Notes in Computer Science},
  site =                {Grenoble, France},
  title =               {{Elastic ChatNoir: Search Engine for the ClueWeb and the Common Crawl}},
  year =                2018
}

@article{C4,
  author  = {Colin Raffel and Noam Shazeer and Adam Roberts and Katherine Lee and Sharan Narang and Michael Matena and Yanqi Zhou and Wei Li and Peter J. Liu},
  title   = {Exploring the Limits of Transfer Learning with a Unified Text-to-Text Transformer},
  journal = {Journal of Machine Learning Research},
  year    = {2020},
  volume  = {21},
  number  = {140},
  pages   = {1--67},
  url     = {http://jmlr.org/papers/v21/20-074.html}
}

@inproceedings{barbaresi2021trafilatura,
  title={Trafilatura: A web scraping library and command-line tool for text discovery and extraction},
  author={Barbaresi, Adrien},
  booktitle={Proceedings of the 59th Annual Meeting of the Association for Computational Linguistics and the 11th International Joint Conference on Natural Language Processing: System Demonstrations},
  pages={122--131},
  year={2021}
}

@InProceedings{survey,
  author =    {Janek Bevendorff and Sanket Gupta and Johannes Kiesel and Benno Stein},
  booktitle = {46th International ACM SIGIR Conference on Research and Development in Information Retrieval (SIGIR 2023)},
  publisher = {ACM},
  site =      {Taipei, Taiwan},
  title =     {{An Empirical Comparison of Web Content Extraction Algorithms}},
  year =      2023,
  url =       {https://dl.acm.org/doi/10.1145/3539618.3591920},
  doi =       {10.1145/3539618.3591920}
}

@misc{commoncrawl,
  author       = {{Common Crawl Foundation}}, 
  title        = {Common Crawl: Open-Source Web Crawl Data \& Infrastructure}, 
  howpublished = {\url{https://commoncrawl.org/}}, 
}

@misc{qwen3technicalreport,
      title={Qwen3 Technical Report}, 
      author={Qwen Team},
      year={2025},
      eprint={2505.09388},
      archivePrefix={arXiv},
      primaryClass={cs.CL},
      url={https://arxiv.org/abs/2505.09388}, 
}

@misc{DCLM2024,
  title = {{{DataComp-LM}}: {{In}} Search of the next Generation of Training Sets for Language Models},
  author = {Li, Jeffrey and Fang, Alex and Smyrnis, Georgios and Ivgi, Maor and Jordan, Matt and Gadre, Samir and Bansal, Hritik and Guha, Etash and Keh, Sedrick and Arora, Kushal and Garg, Saurabh and Xin, Rui and Muennighoff, Niklas and Heckel, Reinhard and Mercat, Jean and Chen, Mayee and Gururangan, Suchin and Wortsman, Mitchell and Albalak, Alon and Bitton, Yonatan and Nezhurina, Marianna and Abbas, Amro and Hsieh, Cheng-Yu and Ghosh, Dhruba and Gardner, Josh and Kilian, Maciej and Zhang, Hanlin and Shao, Rulin and Pratt, Sarah and Sanyal, Sunny and Ilharco, Gabriel and Daras, Giannis and Marathe, Kalyani and Gokaslan, Aaron and Zhang, Jieyu and Chandu, Khyathi and Nguyen, Thao and Vasiljevic, Igor and Kakade, Sham and Song, Shuran and Sanghavi, Sujay and Faghri, Fartash and Oh, Sewoong and Zettlemoyer, Luke and Lo, Kyle and {El-Nouby}, Alaaeldin and Pouransari, Hadi and Toshev, Alexander and Wang, Stephanie and Groeneveld, Dirk and Soldaini, Luca and Koh, Pang Wei and Jitsev, Jenia and Kollar, Thomas and Dimakis, Alexandros G. and Carmon, Yair and Dave, Achal and Schmidt, Ludwig and Shankar, Vaishaal},
  year = {2024},
  number = {arXiv:2406.11794},
  eprint = {2406.11794},
  primaryclass = {cs},
  urldate = {2024-06-27},
  archiveprefix = {arXiv}
}

@article{dubey2024llama,
  title={The llama 3 herd of models},
  author={Dubey, Abhimanyu and Jauhri, Abhinav and Pandey, Abhinav and Kadian, Abhishek and Al-Dahle, Ahmad and Letman, Aiesha and Mathur, Akhil and Schelten, Alan and Yang, Amy and Fan, Angela and others},
  journal={arXiv preprint arXiv:2407.21783},
  year={2024}
}

@misc{eval-harness,
  author       = {Gao, Leo and Tow, Jonathan and Abbasi, Baber and Biderman, Stella and Black, Sid and DiPofi, Anthony and Foster, Charles and Golding, Laurence and Hsu, Jeffrey and Le Noac'h, Alain and Li, Haonan and McDonell, Kyle and Muennighoff, Niklas and Ociepa, Chris and Phang, Jason and Reynolds, Laria and Schoelkopf, Hailey and Skowron, Aviya and Sutawika, Lintang and Tang, Eric and Thite, Anish and Wang, Ben and Wang, Kevin and Zou, Andy},
  title        = {A framework for few-shot language model evaluation},
  month        = 12,
  year         = 2023,
  publisher    = {Zenodo},
  version      = {v0.4.0},
  doi          = {10.5281/zenodo.10256836},
  url          = {https://zenodo.org/records/10256836}
}

@article{gopher,
  title={Scaling language models: Methods, analysis \& insights from training gopher},
  author={Rae, Jack W and Borgeaud, Sebastian and Cai, Trevor and Millican, Katie and Hoffmann, Jordan and Song, Francis and Aslanides, John and Henderson, Sarah and Ring, Roman and Young, Susannah and others},
  journal={arXiv preprint arXiv:2112.11446},
  year={2021}
}

@article{refinedweb,
  title={The RefinedWeb dataset for Falcon LLM: outperforming curated corpora with web data, and web data only},
  author={Penedo, Guilherme and Malartic, Quentin and Hesslow, Daniel and Cojocaru, Ruxandra and Cappelli, Alessandro and Alobeidli, Hamza and Pannier, Baptiste and Almazrouei, Ebtesam and Launay, Julien},
  journal={arXiv preprint arXiv:2306.01116},
  year={2023}
}

@article{dolma,
  title={Dolma: An open corpus of three trillion tokens for language model pretraining research},
  author={Soldaini, Luca and Kinney, Rodney and Bhagia, Akshita and Schwenk, Dustin and Atkinson, David and Authur, Russell and Bogin, Ben and Chandu, Khyathi and Dumas, Jennifer and Elazar, Yanai and others},
  journal={arXiv preprint arXiv:2402.00159},
  year={2024}
}

@article{fineweb,
  title={The FineWeb Datasets: Decanting the Web for the Finest Text Data at Scale},
  author={Penedo, Guilherme and Kydl{\'\i}{\v{c}}ek, Hynek and Lozhkov, Anton and Mitchell, Margaret and Raffel, Colin and Von Werra, Leandro and Wolf, Thomas and others},
  journal={arXiv preprint arXiv:2406.17557},
  year={2024}
}

@inproceedings{winogrande,
  title={WinoGrande: An Adversarial Winograd Schema Challenge at Scale},
  author={Sakaguchi, Keisuke and Le Bras, Ronan and Bhagavatula, Chandra and Choi, Yejin},
  booktitle={Proceedings of the AAAI Conference on Artificial Intelligence},
  volume={34},
  pages={8732--8740},
  year={2020}
}

@inproceedings{commonsenseqa,
    title = "{C}ommonsense{QA}: A Question Answering Challenge Targeting Commonsense Knowledge",
    author = "Talmor, Alon  and
      Herzig, Jonathan  and
      Lourie, Nicholas  and
      Berant, Jonathan",
    booktitle = "Proceedings of the 2019 Conference of the North {A}merican Chapter of the Association for Computational Linguistics: Human Language Technologies, Volume 1 (Long and Short Papers)",
    month = jun,
    year = "2019",
    address = "Minneapolis, Minnesota",
    publisher = "Association for Computational Linguistics",
    url = "https://aclanthology.org/N19-1421",
    doi = "10.18653/v1/N19-1421",
    pages = "4149--4158",
}

@inproceedings{openbookqa,
    title = "Can a Suit of Armor Conduct Electricity? A New Dataset for Open Book Question Answering",
    author = "Mihaylov, Todor  and
      Clark, Peter  and
      Khot, Tushar  and
      Sabharwal, Ashish",
    booktitle = "Proceedings of the 2018 Conference on Empirical Methods in Natural Language Processing",
    year = "2018",
    address = "Brussels, Belgium",
    publisher = "Association for Computational Linguistics",
    url = "https://aclanthology.org/D18-1260",
    doi = "10.18653/v1/D18-1260",
    pages = "2381--2391",
}

@article{arc,
  title={Think you have solved question answering? try arc, the ai2 reasoning challenge},
  author={Clark, Peter and Cowhey, Isaac and Etzioni, Oren and Khot, Tushar and Sabharwal, Ashish and Schoenick, Carissa and Tafjord, Oyvind},
  journal={arXiv preprint arXiv:1803.05457},
  year={2018}
}

@inproceedings{sciq,
    title = "Crowdsourcing Multiple Choice Science Questions",
    author = "Welbl, Johannes  and
      Liu, Nelson F.  and
      Gardner, Matt",
    booktitle = "Proceedings of the 3rd Workshop on Noisy User-generated Text",
    month = sep,
    year = "2017",
    address = "Copenhagen, Denmark",
    publisher = "Association for Computational Linguistics",
    url = "https://aclanthology.org/W17-4413",
    doi = "10.18653/v1/W17-4413",
    pages = "94--106"
 }

@inproceedings{hellaswag,
    title = "{H}ella{S}wag: Can a Machine Really Finish Your Sentence?",
    author = "Zellers, Rowan  and
      Holtzman, Ari  and
      Bisk, Yonatan  and
      Farhadi, Ali  and
      Choi, Yejin",
    editor = "Korhonen, Anna  and
      Traum, David  and
      M{\`a}rquez, Llu{\'\i}s",
    booktitle = "Proceedings of the 57th Annual Meeting of the Association for Computational Linguistics",
    month = jul,
    year = "2019",
    address = "Florence, Italy",
    publisher = "Association for Computational Linguistics",
    url = "https://aclanthology.org/P19-1472",
    doi = "10.18653/v1/P19-1472",
    pages = "4791--4800"
 }

@inproceedings{bisk2020piqa,
  title={Piqa: Reasoning about physical commonsense in natural language},
  author={Bisk, Yonatan and Zellers, Rowan and Gao, Jianfeng and Choi, Yejin and others},
  booktitle={Proceedings of the AAAI conference on artificial intelligence},
  volume={34},
  pages={7432--7439},
  year={2020}
}

@article{siqa,
  title={Socialiqa: Commonsense reasoning about social interactions},
  author={Sap, Maarten and Rashkin, Hannah and Chen, Derek and LeBras, Ronan and Choi, Yejin},
  journal={arXiv preprint arXiv:1904.09728},
  year={2019}
}

@article{reddy2019coqa,
  title={Coqa: A conversational question answering challenge},
  author={Reddy, Siva and Chen, Danqi and Manning, Christopher D},
  journal={Transactions of the Association for Computational Linguistics},
  volume={7},
  pages={249--266},
  year={2019},
  publisher={MIT Press One Rogers Street, Cambridge, MA 02142-1209, USA journals-info~…}
}

@inproceedings{boolq,
  title={BoolQ: Exploring the Surprising Difficulty of Natural Yes/No Questions},
  author={Clark, Christopher and Lee, Kenton and Chang, Ming-Wei and Kwiatkowski, Tom and Collins, Michael and Toutanova, Kristina},
  booktitle={Proceedings of the 2019 Conference of the North American Chapter of the Association for Computational Linguistics: Human Language Technologies, Volume 1 (Long and Short Papers)},
  pages={2924--2936},
  year={2019}
}

@inproceedings{mmlu,
    title={Measuring Massive Multitask Language Understanding},
    author={Dan Hendrycks and Collin Burns and Steven Basart and Andy Zou and Mantas Mazeika and Dawn Song and Jacob Steinhardt},
    booktitle={International Conference on Learning Representations},
    year={2021},
    url={https://openreview.net/forum?id=d7KBjmI3GmQ}
}

@inproceedings{paperno2016lambada,
  title={The LAMBADA dataset: Word prediction requiring a broad discourse context},
  author={Paperno, D and Kruszewski, G and Lazaridou, A and Pham, QN and Bernardi, Raffaella and Pezzelle, S and Baroni, M and Boleda, G and Fern{\'a}ndez, R},
  booktitle={54th Annual Meeting of the Association for Computational Linguistics, ACL 2016-Long Papers},
  volume={3},
  pages={1525--1534},
  year={2016},
  organization={Association for Computational Linguistics (ACL)}
}

@inproceedings{gordon2012semeval,
  title={SemEval-2012 task 7: Choice of plausible alternatives: An evaluation of commonsense causal reasoning},
  author={Gordon, Andrew and Kozareva, Zornitsa and Roemmele, Melissa},
  booktitle={* SEM 2012: The First Joint Conference on Lexical and Computational Semantics--Volume 1: Proceedings of the main conference and the shared task, and Volume 2: Proceedings of the Sixth International Workshop on Semantic Evaluation (SemEval 2012)},
  pages={394--398},
  year={2012}
}

@misc{hoffmann2022training,
  title={Training Compute-Optimal Large Language Models},
  author={Jordan Hoffmann and Sebastian Borgeaud and Arthur Mensch and Elena Buchatskaya and Trevor Cai and Eliza Rutherford and Diego de Las Casas and Lisa Anne Hendricks and Johannes Welbl and Aidan Clark and Tom Hennigan and Eric Noland and Katie Millican and George van den Driessche and Bogdan Damoc and Aurelia Guy and Simon Osindero and Karen Simonyan and Erich Elsen and Jack W. Rae and Oriol Vinyals and Laurent Sifre},
  year={2022},
  eprint={2203.15556},
  archivePrefix={arXiv},
  primaryClass={cs.CL}
}

@article{llama,
  title={Llama: Open and efficient foundation language models},
  author={Touvron, Hugo and Lavril, Thibaut and Izacard, Gautier and Martinet, Xavier and Lachaux, Marie-Anne and Lacroix, Timoth{\'e}e and Rozi{\`e}re, Baptiste and Goyal, Naman and Hambro, Eric and Azhar, Faisal and others},
  journal={arXiv preprint arXiv:2302.13971},
  year={2023}
}

@article{touvron2023llama2,
  title={Llama 2: Open foundation and fine-tuned chat models},
  author={Touvron, Hugo and Martin, Louis and Stone, Kevin and Albert, Peter and Almahairi, Amjad and Babaei, Yasmine and Bashlykov, Nikolay and Batra, Soumya and Bhargava, Prajjwal and Bhosale, Shruti and others},
  journal={arXiv preprint arXiv:2307.09288},
  year={2023}
}

@article{nemotron2024,
  title={Nemotron-CC: Transforming Common Crawl into a Refined Long-Horizon Pretraining Dataset},
  author={Su, Dan and Kong, Kezhi and Lin, Ying and Jennings, Joseph and Norick, Brandon and Kliegl, Markus and Patwary, Mostofa and Shoeybi, Mohammad and Catanzaro, Bryan},
  journal={arXiv preprint},
  year={2024}
}

@misc{dingo,
  title={Dingo: A Comprehensive AI Data Quality Evaluation Tool for Large Models},
  author={Dingo Contributors},
  howpublished={\url{https://github.com/MigoXLab/dingo}},
  year={2024}
}
\bibliographystyle{plainnat}
\setcitestyle{numbers}

% ------------------------------------------------------------------------------------------
\clearpage
\newpage
\beginappendix

\section{MinerU-HTML Details}
\subsection{MinerU-HTML-Classifier}
\label{appdx:MinerU-HTML_classifier}
\subsubsection{Training Data Construction}
To train MinerU-HTML, we construct a large-scale training dataset designed to capture the diversity of the modern web. The dataset is curated through a three-stage pipeline that ensures variety in page layout, language, and document format.

First, we sample structurally diverse pages from Common Crawl by clustering pages based on DOM tree features using DBSCAN, yielding approximately 40 million layout-distinct candidates. Second, we apply language identification and format classification to obtain a balanced 1 million page subset covering multiple languages and web formats. Third, we generate block-level annotations by processing pages through MinerU-HTML's simplification algorithm and using a large language model to label each block. After filtering, we obtain our final training dataset of 870,945 annotated samples.
\subsubsection{Training MinerU-HTML-Classifier}
We employ the Qwen3-0.6B \citep{qwen3technicalreport} model as our base model, which is the smallest model in the Qwen3 series, featuring a 32K context window and support for over 100 languages. Supervised fine-tuning is performed on the full set of 870K samples for a fixed total of 4 epochs.

\subsection{Document Formatting}
\label{appdx:content_list2markdown}
Please see the conversion from content list to Markdown in Figure~\ref{fig:content_list2markdown}.

\begin{figure}[H]
  \centering
  \begin{lstlisting}[language=json, basicstyle=\small\ttfamily, frame=single, numbers=left, breaklines=true]
  [ // content list start
      { // node 1
        "type": "title",
        "bbox": null,
        "raw_content": null,
        "content": {
          "title_content": "Future Trends of Large Models",
          "level": 1
        }
      },
  ]
  \end{lstlisting}
  \caption{One title element of the content list. }
  \label{fig:content_list_sample}
\end{figure}

\begin{figure}[H]
  \centering
  \begin{lstlisting}[language=Python, basicstyle=\small\ttfamily, frame=single, numbers=left, breaklines=true]
markdown = []
for page in content_list:
    for element in page:
        if element['type'] == 'title':
            title_md = '#' * element['content']['level'] + element['content']['title_content']
            markdown.append(title_md)
        elif element['type'] == 'code':
            # render code block
            ...
    
markdown_content = '\n\n'.join(markdown)
\end{lstlisting}
  \caption{Pseudocode of the content list to Markdown conversion.}
  \label{fig:content_list2markdown}
\end{figure}

\begin{figure}[H]
  \centering
  \begin{lstlisting}[language=json, basicstyle=\small\ttfamily, frame=single, numbers=left, breaklines=true]
{
  "track_id": "XXXX",
  "html": "<html><body><h1 cc-select=True>Hello world!</h1><aside>advertisement</aside></body></html>",
  "main_html": "<html><body><h1>Hello world!</h1></body></html>",
  "convert_main_content": "# Hello world!",
  "meta": {
    "language": "en",
    "style": "Normal",
    "level": "easy",
    "table": "without",
    "code": "without",
    "equation": "without"
  }
}
  \end{lstlisting}
  \caption{An example data from WebMainBench. It includes the raw source, the ground-truth Main-HTML, its Markdown conversion, and a rich set of metadata for fine-grained analysis.}
  \label{fig:benchmark_example}
\end{figure}

\clearpage
\section{Evaluation of MinerU-HTML}

\subsection{Benchmark Construction}
\label{appdx:benchmark_construction}
\textbf{Data Sampling.} WebMainBench is constructed using a hybrid sampling strategy to ensure both broad representation and relevance. 90\% of the samples are randomly drawn from the Common Crawl dataset to cover the long-tail web, while the remaining 10\% are sampled from a list of top-ranking websites (Chinaz Alexa\footnote{https://malexa.chinaz.com/}) to include popular, professionally designed pages. The final benchmark is highly diverse, containing pages from 5,434 unique top-level and 5,904 unique second-level domains.

\textbf{Annotation Rules.} To address the ambiguity in defining "main content" for unconventional layouts, we establish two core annotation principles. First, \textbf{Contextual Integrity} dictates that content integral to the main article—such as a table of contents, abstract, or reference list—is included. Conversely, contextually independent elements like "related articles" sidebars or copyright footers are excluded. Second, the main content is defined as \textbf{Human-Generated Content}, including article bodies, user comments, and Q\&A posts, while associated auto-generated metadata like usernames and timestamps are excluded.

\textbf{Annotation Process.} The annotation for each page followed a rigorous three-stage process using a custom-built tool (see Appendix Figure~\ref{fig:snapshot}) that allowed for tag-level granularity. The process involved: (1) an initial pass by one annotator, (2) a review and correction pass by a second annotator, and (3) a final quality assurance check by a senior inspector, who made the final adjudication to resolve any discrepancies. Pages uninterpretable due to rendering issues were discarded.

\textbf{Metadata Annotation.} To enable detailed, fine-grained analysis, we annotate each page with a rich set of metadata. This includes \textbf{Language}, identified by GPT-5 and labeled as \texttt{en} (English) or \texttt{non\_en} (other), and \textbf{Style}, classified by GPT-5 as \texttt{Conversational} for pages with user-generated content or \texttt{Normal} otherwise. We also develop a quantitative \textbf{Difficulty Level}, determined by an \texttt{overall\_complexity\_score} calculated for each page. To compute this score, we first measure four distinct metrics: \textit{DOM structural complexity} (based on tree depth and width), \textit{text distribution sparsity} (transitions between text/non-text nodes), \textit{content-type diversity} (a count of rich content types), and \textit{link density} (the ratio of hyperlinked text). These four values are individually normalized, and their weighted sum produces the final score. Based on the distribution of this \texttt{overall\_complexity\_score} across the benchmark, we then categorize pages into \texttt{simple}, \texttt{medium}, and \texttt{hard} using the 30th and 70th percentiles as dynamic thresholds. Finally, we add \textbf{Rich Content Tags }to identify the presence of tables (\texttt{<table>}), code blocks (\texttt{<code>}), and mathematical formulas (\texttt{<math>} or LaTeX patterns) using BeautifulSoup.

\begin{table}[H]
\centering
\caption{Details of Web Content Extraction Datasets}
\small
\setlength{\tabcolsep}{5pt}
\begin{tabularx}{\textwidth}{@{}p{2cm}rX@{}}
\toprule
\textbf{Dataset} & \textbf{Pages} & \textbf{Source \& Characteristics} \\
\midrule
CleanEval & 738 & De-facto standard dataset from 2007 shared task combining development and evaluation sets of English web pages with basic structural markup ground truth \\
\midrule
CleanPortalEval & 71 & Extension of CleanEval featuring multi-page samples from 4 major news domains  \\
\midrule
CETD & 700 & Created for density-based extractor evaluation across 6 domains \\
\midrule
Dragnet & 1,379 & Combined sources from popular RSS feeds, 23 major news sites, 178 Technorati blogs, plus CETR and CleanEval conversions \\
\midrule
L3S-GN1 & 621 & Created by BoilerPipe authors with unique HTML annotation using span-wrapped CSS classes for 5-level content relevance \\
\midrule
Google-Trends-2017 & 180 & Dataset created for BoilerNet neural network training featuring binary CSS class annotations on DOM leaf nodes to distinguish content from boilerplate \\
\midrule
Readability & 115 & Mozilla reader mode test suite with original and simplified HTML for evaluation \\
\midrule
Scrapinghub & 181 & Created by Zyte for benchmarking proprietary extraction services \\
\bottomrule
\end{tabularx}
\label{tab:datasets}
\end{table}

\begin{table}[H]
  \centering
  \caption{Distribution of Structured Elements in WebMainBench-Structured (N=545 pages). Element counts and subtype breakdowns for formulas, code blocks, and tables.}
  \label{tab:elements_distribution}
  \begin{tabular}{llll}
  \toprule
  \textbf{Element Type}             & \textbf{Count}                & \textbf{SubType}   & \textbf{SubType Count} \\
  \midrule
  \multirow{3}{*}{Formula} & \multirow{3}{*}{257} & Inline             & 198                    \\
  &                               & Interline          & 9                      \\
  &                               & Inline + Interline & 50                     \\
  \midrule
  \multirow{3}{*}{Code}             & \multirow{3}{*}{127}          & Inline             & 30                     \\
  &                               & Interline          & 73                     \\
  &                               & Inline + Interline & 24                     \\
  \midrule
  \multirow{3}{*}{Table}            & \multirow{3}{*}{179}          & Data               & 151                    \\
                                    &                               & Layout             & 6                      \\
                                    &                               & Data + Layout      & 22                     \\
  \bottomrule
  \end{tabular}
\end{table}

\section{Pretraining Details}
\label{appdx:pretrain_detail}
The architecture of pretrained models in this work is as follows: 
Number of Layers: 24, Hidden Size: 2048, Intermediate Size: 5504, Number of Attention Heads: 16, Number of KV Heads: 16, Head Dimension=128, RoPE Theta: 1000000.0, Number of Total Parameters: 1,525,516,288, 

The pretraining step is as follows: Global Batch Size: 64, Context Length: 4,096, Training Steps: 240,000, Consumed Tokens: 62B.

We employ the Qwen3 tokenizer \citep{qwen3technicalreport} with a vocabulary size of 151,936.
The learning rate is set to $1 \times 10^{-4}$, and the AdamW optimizer is employed with hyperparameters ($ Weight\_decay=0.01, \beta_1=0.9, \beta_2=0.95, \epsilon=10^{-8}$).

\section{Evaluation Details}
\label{appdx:evaluation}
The number of randomly selected demonstrations for few-shot in-context learning for each task is listed in Table~\ref{tab:icl}.

\begin{table}[H]
\centering
\small
\resizebox{\textwidth}{!}{%
\begin{tabular}{@{}lcccccccccccccc@{}}
\toprule
\textbf{Task} & \textbf{ARC} & \textbf{BoolQ} & \textbf{CSQA} & \textbf{MMLU} & \textbf{SciQ} & \textbf{COPA} & \textbf{HellaSwag} & \textbf{PIQA} & \textbf{SIQA} & \textbf{WinoGrande} & \textbf{CoQA} & \textbf{LAMBADA} & \textbf{OpenbookQA} \\
\midrule
\textbf{\# Shots} & 10 & 10 & 10 & 5 & 5 & 0 & 10 & 10 & 5 & 0 & 0 & 0 & 0 \\
\bottomrule
\end{tabular}%
}
\caption{Number of demonstrations in in-context learning (few-shot) used for each downstream task. CSQA = CommonsenseQA.}
\label{tab:icl}
\end{table}

The full performance results are shown in Table~\ref{tab:Perf_2methods_dynamic} and Table~\ref{tab:Perf_4methods_last_ckpt}.
\begin{table}[H]
  \centering
  \caption{Average accuracy (\%) across all 13 benchmarks during training. Models trained on AICC and TfCC are evaluated at 15 checkpoints from 4B to 63B tokens. AICC consistently maintains superior performance throughout training.}
  \label{tab:Perf_2methods_dynamic}
  \resizebox{\textwidth}{!}{%
  \begin{tabular}{lrrrrrrrrrrrrrrr}
  \toprule
  \textbf{Method} & \textbf{4.19} & \textbf{8.39} & \textbf{12.58} & \textbf{16.78} & \textbf{20.97} & \textbf{25.17} & \textbf{29.36} & \textbf{33.55} & \textbf{37.75} & \textbf{41.94} & \textbf{46.14} & \textbf{50.33} & \textbf{54.53} & \textbf{58.72} & \textbf{62.91} \\
  \midrule
  AICC    & 41.12 & 44.47 & 45.25 & 47.21 & 47.79 & 47.85 & 48.18 & 48.55 & 49.21 & 49.36 & 49.43 & 50.43 & 49.61 & 50.64 & \textbf{50.82} \\
  TfCC            & 40.68 & 43.03 & 44.62 & 45.34 & 46.75 & 47.72 & 47.85 & 48.49 & 48.51 & 49.09 & 49.19 & 49.49 & 49.65 & 49.82 & 49.74 \\
  \bottomrule
  \end{tabular}%
  }
\end{table}

\begin{table}[H]
  \centering
  \caption{Performance breakdown by task category at the final checkpoint (63B tokens). Accuracy scores (\%) comparing AICC, TfCC, FineWeb, and RefinedWeb across three task categories. AICC achieves the best performance on General Knowledge and Reading Comprehension tasks, while FineWeb performs best on Reasoning tasks. General Knowl. = General Knowledge, Reading Compr. = Reading Comprehension.}
  \label{tab:Perf_4methods_last_ckpt}
  \begin{tabular}{lcccc}
  \toprule
  \textbf{Task Category} & \textbf{AICC} & \textbf{TfCC} & \textbf{FineWeb} & \textbf{RefinedWeb} \\
  \midrule
    General Knowl. & 47.54 & 45.61 & 46.86   & 44.57      \\
    Reasoning      & 59.83 & 59.34 & 60.69   & 59.43      \\
    Reading Compr. & 42.37 & 42.02 & 36.68   & 41.10      \\
  \bottomrule
  \end{tabular}
\end{table}

\clearpage
\section{Structured Element Extraction Case Study}

\label{appdx:annotation_tool}

\begin{figure}
    \centering
    \includegraphics[width=1\linewidth]{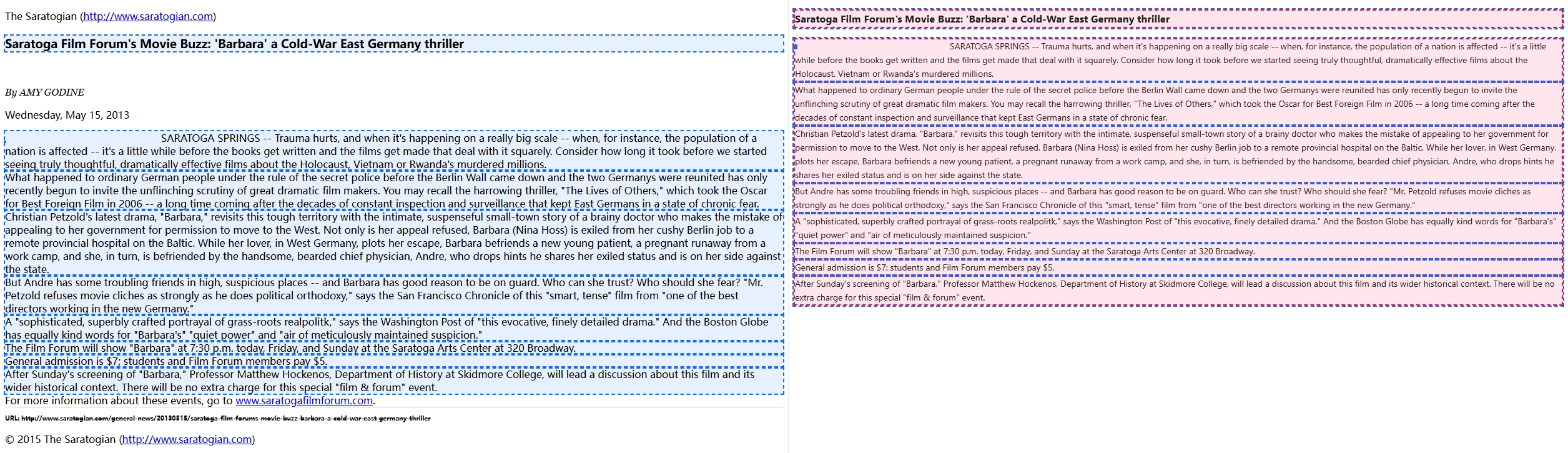}
    \caption{Screenshot of the web page annotation tool. The main content selection is highlighted in blue on the left, with a real-time preview on the right.}
    \label{fig:snapshot}
\end{figure}

\clearpage
\section{AICC-TfCC pairwise document comparison}
\label{appdx:AICC-TfCC_comparison}

\begin{figure}[htbp]
  \centering
  \begin{lstlisting}[
  basicstyle=\scriptsize\ttfamily,
  frame=single,
  breaklines=true,
  backgroundcolor=\color{gray!5}
]
You are a professional HTML content extraction evaluator, skilled in analyzing 
the conversion quality between HTML code and Markdown text. I will provide three 
pieces of content:

1. **Original HTML Code**: The complete HTML structure of the webpage.
2. **Tool A's Extracted Markdown**: Markdown text extracted from HTML, suitable 
   for LLM training.
3. **Tool B's Extracted Markdown**: Markdown text extracted from HTML, suitable 
   for LLM training.

Note: The order of Tool A and Tool B is not fixed. Do not favor either tool 
based on order; evaluate objectively based on actual conversion quality.

Your Task:
1. Compare both Markdown extractions against the HTML code. Strictly check 
   extraction effectiveness for the following 8 module types:

**HTML Element Identification:**
- `code`: Code blocks (<pre>, <code> tags)
- `math`: Mathematical formulas (MathJax, MathML, LaTeX)
- `table`: Tables (<table> tags)
- `image`: Images (<img> tags)
- `list`: Ordered/unordered lists (<ul>, <ol> tags)
- `title`: Headings (<h1>-<h6> tags)
- `paragraph`: Paragraph text (<p>, <div> containers)
- `other`: Other visible content not covered above

**Markdown Element Statistics:**
- Code blocks: ```...``` or indented code
- Formulas: $...$ $$...$$ \(...\) \[...\]
- Tables: |...| format
- Images: ![](...) format
- Lists: -, *, 1. markers
- Headings: #, ## markers
- Paragraphs: Plain text blocks

2. **Scoring Rules**: Evaluate which tool has better extraction quality.
   - **Extraction Completeness**: Check if key content (code, tables, images, 
     lists) is fully extracted.
   - **Format Accuracy**: Verify correct Markdown formatting (code indentation, 
     table alignment, image links).
   - **Semantic Coherence**: Ensure logical flow and heading hierarchy are 
     preserved.

3. **Issue Feedback**: Strictly identify problems by the 8 module types above; 
   return empty list if no issues.

4. **Return Result**: JSON format with 3 fields: score, name, reason.
   - `score`: 1 if Tool A is better, 2 if Tool B is better.
   - `name`: Must be one of the 8 module types, selecting the module with 
     greatest difference.
   - `reason`: Objective description of performance differences in that module.

Example Output:
{
  "score": [1|2],
  "name": "[module_type]",
  "reason": "[objective description of differences]"
}
\end{lstlisting}
\caption{Rating prompt used for automated pairwise document comparison. The prompt instructs the judge model to compare two HTML extraction tools by evaluating 8 content module types (code, math, table, image, list, title, paragraph, other) and return a structured JSON score. (Part 1/2)}
\end{figure}
\begin{figure}[t]\ContinuedFloat
\centering
\begin{lstlisting}[
  basicstyle=\scriptsize\ttfamily,
  frame=single,
  breaklines=true,
  backgroundcolor=\color{gray!5}
]
**Important Notes**:
1. Only use predefined module categories.
2. Focus on structured content (code, tables, formulas, images) conversion.
3. For paragraphs, check text coherence and semantic completeness.

### Original HTML Code:
```html
{}
```

### Tool A's Extracted Markdown:
```md
{}
```

### Tool B's Extracted Markdown:
```md
{}
```

Return ONLY one JSON object, no other explanations or analysis!
  \end{lstlisting}
  \caption{Rating prompt used for automated pairwise document comparison. The prompt instructs the judge model to compare two HTML extraction tools by evaluating 8 content module types (code, math, table, image, list, title, paragraph, other) and return a structured JSON score. (Part 2/2)}
  \label{fig:rating_prompt}
\end{figure}

\end{document}